\documentclass{article} 
\usepackage{iclr2027_conference,times}

\usepackage{amsmath,amsfonts,bm}

\def\eqref#1{equation~\ref{#1}}

\def\1{\bm{1}}

\DeclareMathAlphabet{\mathsfit}{\encodingdefault}{\sfdefault}{m}{sl}
\SetMathAlphabet{\mathsfit}{bold}{\encodingdefault}{\sfdefault}{bx}{n}

\usepackage{hyperref}
\usepackage{url}

\usepackage{graphicx}
\usepackage{booktabs}       
\usepackage{amsfonts}       
\usepackage{nicefrac}       
\usepackage{xcolor}         

\usepackage{amsmath}
\usepackage{amssymb}
\usepackage{graphicx}
\usepackage{multirow}
\usepackage{bm}
\usepackage{array}
\usepackage{makecell}
\usepackage{adjustbox}
\usepackage[table]{xcolor}
\usepackage{pifont}
\usepackage{algorithm}
\usepackage{algpseudocode}
\usepackage{enumitem}

\usepackage{cuted}
\usepackage{wrapfig}
\usepackage{caption}

\newcommand{\cmark}{\ding{51}}
\newcommand{\xmark}{\ding{55}}

\title{GeoCR: Learning a Generalist Cloud Removal Prior from Heterogeneous Observations}

\author{
Jeonghyeok Do \quad Munchurl Kim \thanks{Corresponding author.}\\
Korea Advanced Institute of Science and Technology (KAIST)\\
\texttt{\{ehwjdgur0913,mkimee\}@kaist.ac.kr}\\[4pt]
{\small Project Page: \url{https://kaist-viclab.github.io/GeoCR_site/}}
}

\iclrfinalcopy
\begin{document}

\maketitle

\begin{center}
    \vspace{-0.8cm}
    \includegraphics[width=0.8\linewidth]{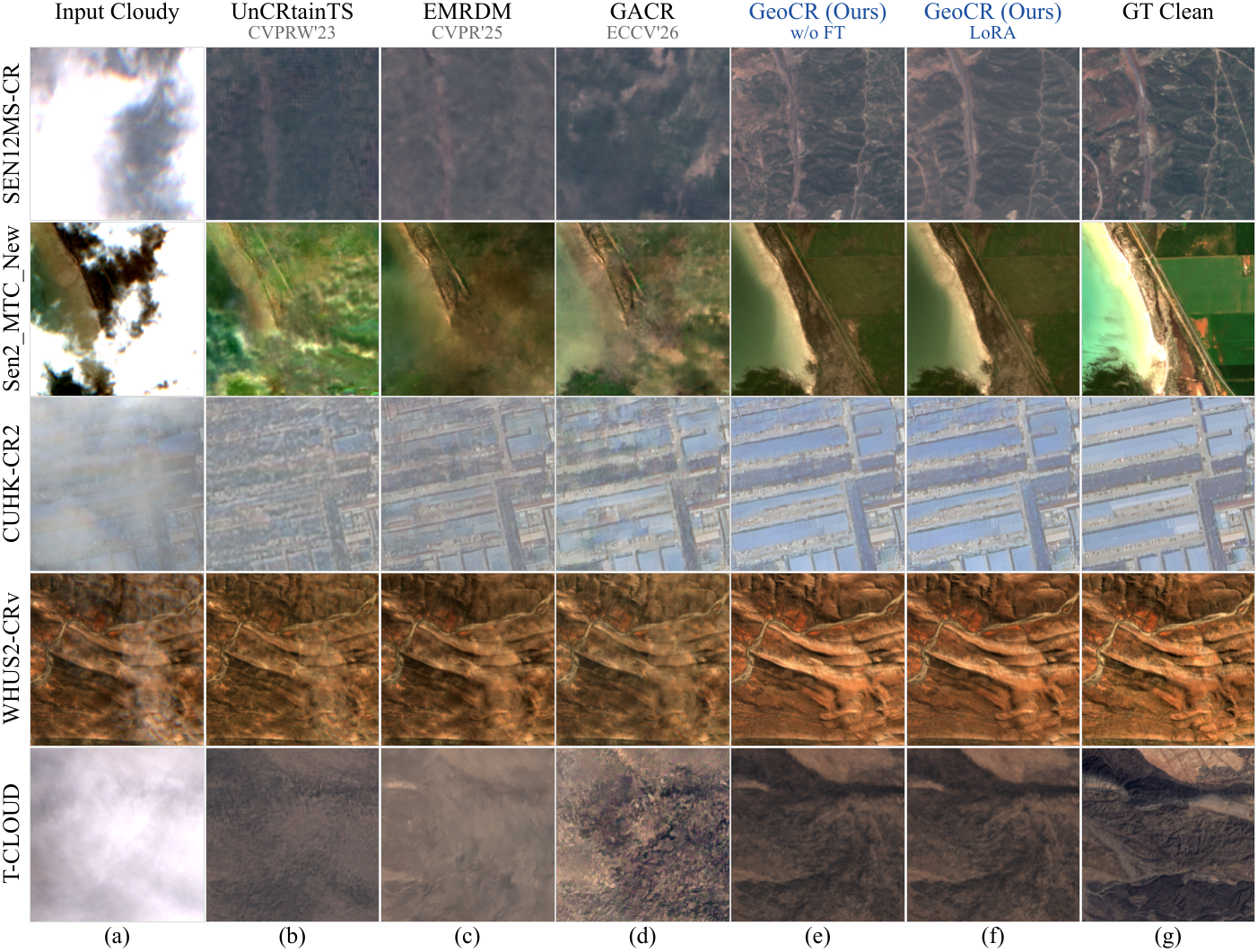}
    \vspace{-0.3cm}
    \captionof{figure}{\textbf{Qualitative comparison on cloud-removal benchmarks.}
    (a) Input cloudy, (b) UnCRtainTS,
    (c) EMRDM, (d) GACR, (e) GeoCR without fine-tuning,
    (f) GeoCR with LoRA adaptation, and (g) ground-truth cloud-free image.}
    \label{fig:first}
\end{center}

\begin{abstract}
Cloud removal methods are typically specialized to individual
datasets and input configurations, limiting reuse across sensors,
spectral bands, and observation settings.
We introduce \textbf{GeoCR}, a generalist model that unifies RGB-only-based CR and multispectral-based CR from single- or multi-temporal cloudy observations, with optional SAR guidance, within a single network.
To accommodate different spectral and sensing domains, compact
input and output stems extend a pretrained RGB autoencoder while
keeping its encoder and decoder trunks frozen.
This shared latent interface enables a single flow transformer
to jointly model clean RGB and non-RGB latents, conditioned on
separate cloudy-observation streams and optional SAR tokens.
Through joint pretraining on the training splits of ten datasets
comprising 883,331 cloud-free target images, GeoCR learns a 
\textit{shared cloud removal prior} across these heterogeneous configurations.
The same pretrained checkpoint supports direct inference without
dataset-specific fine-tuning and efficient adaptation through
low-rank adaptation (LoRA).
We evaluate GeoCR against general image restoration and cloud
removal methods on test splits of the contributing
datasets under full-band and RGB-only settings. GeoCR achieves
the best FID and DISTS on full-band SEN12MS-CR and
Sen2\_MTC\_New and RGB-only CUHK-CR2, outperforming existing models and demonstrating the
effectiveness of a reusable generative model across diverse settings.
\end{abstract}    
\section{Introduction}
\label{sec:introduction}

\begin{wrapfigure}{r}{0.45\columnwidth}
  \centering
  \vspace{-0.4cm}
  \includegraphics[width=0.45\columnwidth]{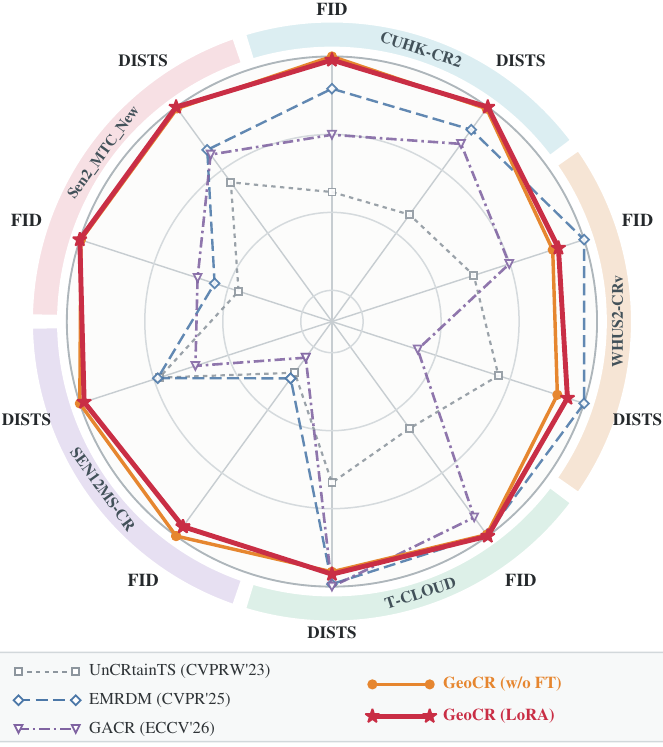}
  \vspace{-0.4cm}
  \caption{\textbf{Cross-dataset comparison of cloud removal (CR) methods.}
  We report FID and DISTS on five cloud removal benchmarks, independently normalized
  for each dataset--metric pair as \(100 \times \text{best}/\text{value}\).}
  \vspace{-0.4cm}
  \label{fig:teaser}
\end{wrapfigure}

Optical satellite imagery provides rich spatial and spectral
information for monitoring the Earth's surface.
Clouds and haze obscure these observations and interrupt
the temporal coverage needed for consistent
analysis~\citep{ebel2023uncrtaints}.
Cloud removal (CR) reconstructs cloud-free optical imagery
using visible content and complementary observations.
The available evidence varies substantially: some applications
provide a single cloudy RGB image, whereas others offer
multispectral time series and synthetic aperture radar (SAR)
measurements.
A reusable CR model must therefore reconstruct missing content
while accommodating different spectral bands, temporal coverage,
and sensing modalities.

Recent methods have advanced CR through attention-based
temporal fusion~\citep{ebel2023uncrtaints} and diffusion-based
restoration~\citep{zou2024diffcr,liu2025effective}.
However, they are typically trained separately for individual
datasets and input configurations, limiting reuse across
observation settings.
Large datasets such as AllClear~\citep{zhou2024allclear}
provide opportunities for broader training, but combining
heterogeneous sources requires a model that accommodates
differences in both conditioning observations and output bands.
This raises a central question:
\emph{Can one model learn a shared CR prior that supports RGB-only-based and multispectral-based reconstructions from
single- or multi-temporal observations, with or without SAR?}

To the best of our knowledge, we address \textit{for the first time} this question with \textbf{GeoCR}, a generalist model that brings these observation and reconstruction settings into a single generative framework. Here, generalist refers to one pretrained checkpoint supporting multiple datasets and observation configurations.
We construct a unified corpus from the training splits of ten CR datasets, comprising 883,331 cloud-free target images.
Spanning Sentinel-2, Landsat-8, and high-resolution optical imagery, this corpus supports joint learning across different
spectral, temporal, and SAR configurations.
GeoCR learns a \emph{shared cloud removal prior} from these sources for direct inference and optional dataset-specific adaptation. This unification requires a common representation for
heterogeneous measurements.
Although optical bands differ in their spectral responses, they can share spatial structure, including scene layout and object boundaries; SAR provides complementary structural evidence through a different sensing mechanism.
These properties motivate reusing pretrained spatial features while adapting the interfaces to each measurement domain.
Our \emph{unified embedding framework} connects non-RGB bands
and SAR observations to a pretrained RGB autoencoder through
compact, configuration-specific input and output stems.
The encoder and decoder trunks remain frozen, and the original
RGB path is preserved.
Training only the stems that require adaptation provides a
common latent interface while retaining the pretrained codec.

Building on this interface, we repurpose a pretrained image
flow transformer for conditional CR through flow
matching~\citep{lipman2022flow}.
The generator jointly models clean RGB and available non-RGB
latents using the supplied cloudy observations and optional SAR.
RGB and non-RGB latents are combined within each optical
observation, while different temporal observations and SAR
retain separate conditioning streams.
This design lets one generator combine complementary evidence
across varying input configurations.
After joint pretraining, the same checkpoint supports direct
inference without dataset-specific fine-tuning
(w/o FT) and parameter-efficient adaptation through
LoRA~\citep{hu2021lora}.

We evaluate GeoCR against general image restoration and CR
methods on held-out test splits of the contributing datasets,
covering five primary benchmarks under full-band and RGB-only
settings.
Figure~\ref{fig:teaser} summarizes the comparison, with
GeoCR (w/o FT) using the same checkpoint across all
five benchmarks. GeoCR achieves
the lowest FID and DISTS among the compared methods on
full-band SEN12MS-CR and Sen2\_MTC\_New and RGB-only CUHK-CR2.
These results demonstrate that a shared pretrained model
can provide competitive distributional and perceptual quality
across diverse CR configurations without requiring
dataset-specific optimization for every setting.

Our contributions are summarized as follows:
\begin{itemize}[leftmargin=3.0em]
\item We introduce \textbf{GeoCR}, a generalist model that unifies \textit{firstly} RGB-only-based CR and multispectral-based CR across single- and
multi-temporal observations with optional SAR guidance.

\item We develop a unified embedding and conditioning framework that extends frozen RGB encoder--decoder trunks through compact stems and combines observation-specific conditioning streams within a single flow transformer.

\item We consolidate ten CR datasets comprising 883,331
cloud-free target images for joint pretraining.
Extensive evaluations under full-band and RGB-only settings demonstrate state-of-the-art FID and DISTS on three of five primary benchmarks, \textit{significantly} outperforming existing CR models.
\end{itemize}
\section{Related Work}
\label{sec:related_work}

\paragraph{General image translation and restoration.}
Image translation and restoration provide complementary approaches to recovering images from degraded observations.
Pix2pix~\citep{isola2017image} learns paired mappings through conditional adversarial training, while pix2pixHD~\citep{wang2018high} extends it to high-resolution synthesis with coarse-to-fine generation and multiscale discrimination.
BBDM~\citep{li2023bbdm} formulates image translation as a
Brownian bridge between source and target domains.
HI-Diff~\citep{chen2023hierarchical} generates compact latent
priors with diffusion and integrates them hierarchically
into a regression-based deblurring network.
These methods provide general-purpose baselines for evaluating
adversarial, bridge-based, and diffusion-assisted approaches
to cloud removal.

\paragraph{Cloud removal (CR).}
CR methods exploit temporal redundancy, complementary sensor
observations, and generative priors.
UnCRtainTS~\citep{ebel2023uncrtaints} aggregates optical and
SAR observations through temporal attention while estimating
reconstruction uncertainty.
DiffCR~\citep{zou2024diffcr} develops an efficient conditional
diffusion framework for CR.
IDF-CR~\citep{wang2024idfcr} combines pixel-space restoration
with latent diffusion refinement.
EMRDM~\citep{liu2025effective} establishes mean-reverting
dynamics between cloudy and clear images for single- and
multi-temporal restoration.
ThiefCloud~\citep{zhao2025thiefcloud} learns a cloud-thickness
prior for thin-cloud removal, while
GACR~\citep{wang2026gacr} combines observation-anchored residual
flow with alignment to pretrained visual representations.
Related efforts explore unified restoration, multisensor
tokenization, and spatiotemporal modeling in remote
sensing~\citep{cui2026llars,lehmann2026eovae,zhang2025units}.
Our GeoCR focuses on jointly learning a reusable CR prior across heterogeneous datasets and observation configurations.
Its shared latent interface and observation-specific conditioning streams support RGB-only-based and multispectral-based reconstructions with varying temporal coverage and SAR availability, using one pretrained checkpoint for direct inference and optional LoRA adaptation.

\paragraph{Cloud removal datasets.}
CR datasets cover diverse sensing conditions and reconstruction settings.
RICE~\citep{lin2019rice} provides paired imagery for thin- and thick-cloud removal, while
CUHK-CR~\citep{sui2024diffusion} offers high-resolution RGB--NIR observations.
SEN12MS-CR~\citep{ebel2021multisensor} pairs cloudy and clear multispectral optical imagery with SAR measurements.
Sen2\_MTC\_Old~\citep{sarukkai2020cloud} and
Sen2\_MTC\_New~\citep{huang2022ctgan} provide multiple cloudy observations for temporal reconstruction.
AllClear~\citep{zhou2024allclear} expands geographic coverage and multisensor time series, demonstrating the benefits of larger training sets and complementary observations.
These resources differ in spectral coverage, radiometric processing, spatial resolution, temporal sampling, and SAR availability.
Our GeoCR accommodates these differences through a common representation and conditioning framework, enabling joint pretraining across heterogeneous CR datasets.

\begin{figure}[tbp]
  \centering
  \includegraphics[width=1.0\textwidth]{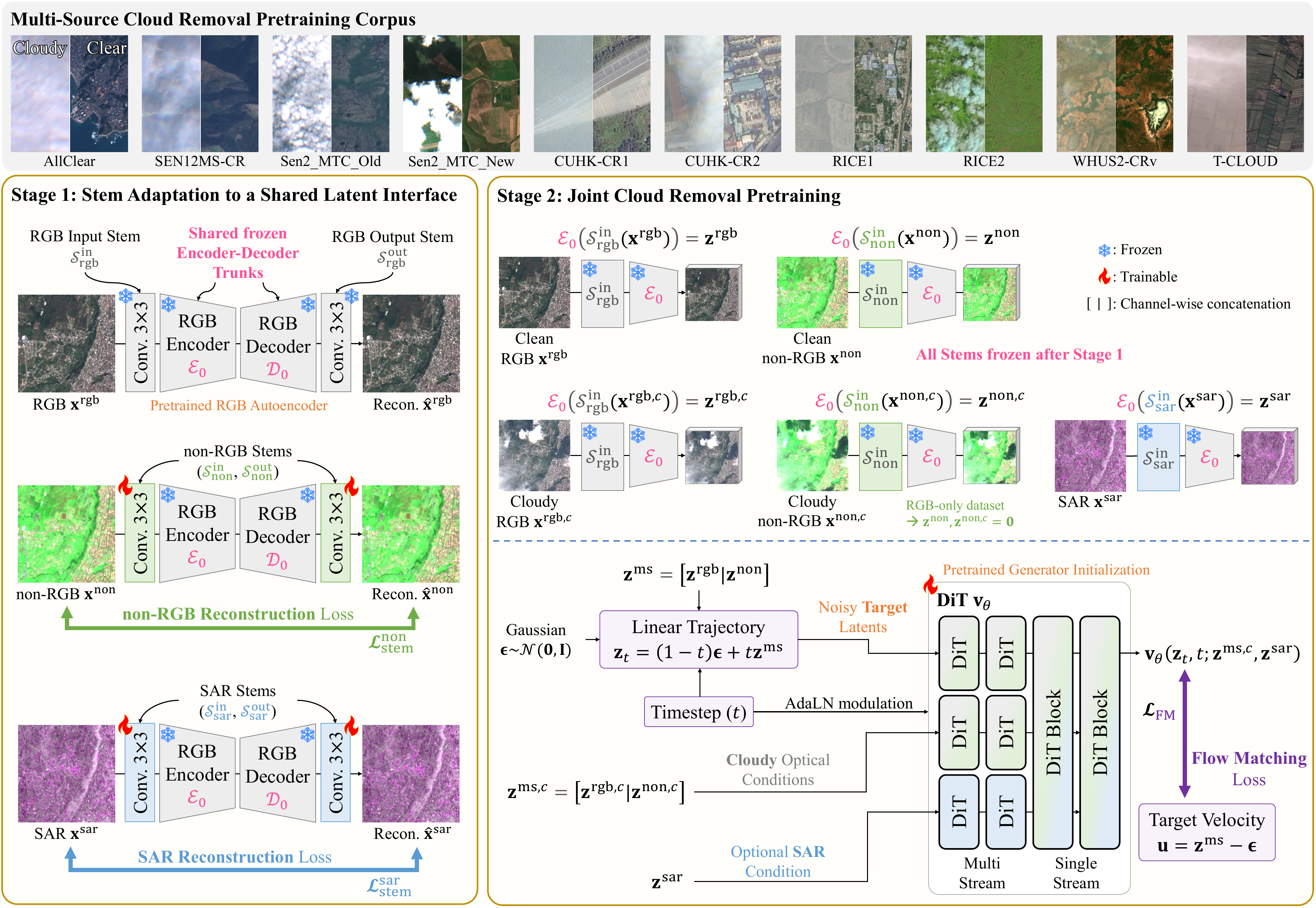}
  \caption{\textbf{GeoCR framework.} A shared latent interface enables joint cloud removal across heterogeneous spectral, temporal, and SAR configurations.}
  \label{fig:framework}
\end{figure}
\section{GeoCR}
\label{sec:method}

\textbf{GeoCR} learns a shared CR prior through two pretraining
stages followed by direct inference or dataset-specific adaptation.
Figure~\ref{fig:framework} illustrates the framework.
Stage~1 trains compact input and output stems for ten-band non-RGB and SAR observations while keeping the shared pretrained encoder--decoder trunks and original RGB path frozen.
Stage~2 learns a shared cloud removal prior through conditional flow matching on the heterogeneous corpus, jointly generating clean RGB and non-RGB latents.
Each cloudy optical observation and the optional SAR input retain separate conditioning streams.
Finally, the same pretrained checkpoint is used directly
(without fine-tuning) or adapted through LoRA~\citep{hu2021lora}.

\paragraph{Problem setting.}
Each training sample contains a cloud-free optical target
$\mathbf{x}=(\mathbf{x}^{\mathrm{rgb}},\mathbf{x}^{\mathrm{non}})$
and $K$ cloudy optical observations
$\{(\mathbf{x}^{\mathrm{rgb,c}}_k,
\mathbf{x}^{\mathrm{non,c}}_k)\}_{k=1}^{K}$,
optionally accompanied by SAR imagery $\mathbf{x}^{\mathrm{sar}}$.
Here, the superscripts `$\mathrm{c}$' and `$\mathrm{non}$', respectively, denote cloudy observations and non-RGB spectral bands that may be absent from the target or conditioning observations.
We use $d$ to index the spectral and radiometric configuration that selects the corresponding non-RGB stem.
For ten-band non-RGB inputs, we distinguish
top-of-atmosphere (TOA) reflectance and
bottom-of-atmosphere (BOA) imagery.

\subsection{Stage 1: Stem Adaptation to a Shared Latent Interface}
\label{sec:stem_adaptation}

\paragraph{Shared pretrained encoder and decoder.}
As shown in Figure~\ref{fig:framework}, different optical bands can share scene layout and spatial boundaries despite differences in their spectral responses.
This motivates sharing pretrained spatial features while adapting the interfaces to different measurement domains.
Let $\mathcal{E}_0$ and $\mathcal{D}_0$ denote the shared encoder
and decoder trunks of a pretrained RGB autoencoder~\citep{flux-2-2025}, excluding
their input and output projections.
For a representation $m$, we define
\begin{equation}
    \mathcal{E}_m
    =\mathcal{E}_0\circ\mathcal{S}^{\mathrm{in}}_m,
    \qquad
    \mathcal{D}_m
    =\mathcal{S}^{\mathrm{out}}_m\circ\mathcal{D}_0,
    \label{eq:geocr_codec}
\end{equation}
where $m\in\{\mathrm{rgb},(\mathrm{non},d),\mathrm{sar}\}$.
The input and output stems,
$\mathcal{S}^{\mathrm{in}}_m$ and $\mathcal{S}^{\mathrm{out}}_m$,
each of which consists of a single convolutional layer.
The RGB stems are the inherited pretrained projections; additional stems are initialized from these projections.
All routes share the same frozen trunks, $\mathcal{E}_0$ and $\mathcal{D}_0$.

\paragraph{Reconstruction-based stem adaptation.}
For input $\mathbf{x}^{m}$, the reconstruction and stem objective are defined as
\begin{equation}
    \widehat{\mathbf{x}}^{m}
    =\mathcal{D}_m\!\left(\mathcal{E}_m(\mathbf{x}^{m})\right),
    \qquad
    \mathcal{L}^{m}_{\mathrm{stem}}
    =\mathbb{E}_{\mathbf{x}^{m}}
    \left[
      \left\|
        \widehat{\mathbf{x}}^{m}-\mathbf{x}^{m}
      \right\|_1
    \right],
    \label{eq:geocr_stem_reconstruction}
\end{equation}
where the $\ell_1$ reconstruction error is measured in the preprocessed input domain.
Only the input and output stems selected for adaptation are optimized while $\mathcal{E}_0$ and $\mathcal{D}_0$ remain fixed.
We train stems for the ten-band non-RGB and dual-polarization SAR configurations, while retaining the original RGB path
without stem training.
Each route reconstructs its own observation, so this stage
learns a representation interface without requiring a cloudy-to-clear
mapping or an RGB reconstruction target for non-RGB measurements.
Fixing the trunks constrains adaptation to their pretrained
feature interfaces while allowing the stems to accommodate
different channel layouts and measurement statistics.
All stems and trunks remain frozen after Stage~1.

\subsection{Stage 2: Joint Cloud Removal Pretraining}
\label{sec:generalist_pretraining}

\paragraph{Optical and SAR latents.}
We encode cloud-free targets and cloudy conditions using the corresponding frozen routes having the following latents in a shared RGB latent space:
\begin{equation}
\begin{aligned}
    \mathbf{z}^{\mathrm{rgb}}
      &=\mathcal{E}_{\mathrm{rgb}}(\mathbf{x}^{\mathrm{rgb}}),
    &\mathbf{z}^{\mathrm{non}}
      &=\mathcal{E}_{\mathrm{non},d}(\mathbf{x}^{\mathrm{non}}),\\
    \mathbf{z}^{\mathrm{rgb,c}}_k
      &=\mathcal{E}_{\mathrm{rgb}}(\mathbf{x}^{\mathrm{rgb,c}}_k),
    &\mathbf{z}^{\mathrm{non,c}}_k
      &=\mathcal{E}_{\mathrm{non},d}(\mathbf{x}^{\mathrm{non,c}}_k).
\end{aligned}
\label{eq:geocr_optical_latents}
\end{equation}
Each available optical component ($\mathrm{rgb}$ or $\mathrm{non}$) is mapped to a latent with the
same channel dimension and spatial grid.
We concatenate the RGB and non-RGB latents along the channel dimension to form joint multispectral (MS) representations of the target and each cloudy observation as:
\begin{equation}
    \mathbf{z}^{\mathrm{ms}}
      =[\mathbf{z}^{\mathrm{rgb}}\mid\mathbf{z}^{\mathrm{non}}],
    \qquad
    \mathbf{z}^{\mathrm{ms,c}}_k
      =[\mathbf{z}^{\mathrm{rgb,c}}_k\mid
        \mathbf{z}^{\mathrm{non,c}}_k],
    \label{eq:geocr_joint_latents}
\end{equation}
where $[\cdot\mid\cdot]$ denotes channel-wise concatenation and `$\mathrm{ms}$' stands for multispectral, combining RGB and non-RGB information.
An unavailable non-RGB component is replaced by a zero latent of the corresponding shape.
This retains a common input layout for RGB-only and multispectral
samples.
When available, SAR is encoded as
$\mathbf{z}^{\mathrm{sar}}
=\mathcal{E}_{\mathrm{sar}}(\mathbf{x}^{\mathrm{sar}})$.
We denote the complete set of available optical and SAR
conditions by $\mathcal{C}$.

\paragraph{Repurposing a pretrained generator.}
We initialize GeoCR from a pretrained image flow
transformer~\citep{flux-2-2025} and condition it on spatial observations instead of language.
The noisy target, cloudy optical conditions, and optional SAR condition are projected into spatial tokens.
Each of the $K$ cloudy observations forms a separate conditioning stream, yielding $1+K$ streams including the target, with one
additional stream when SAR is available.
Multi-stream blocks use separate parameter sets for target, optical-conditioning, and SAR tokens, with parameters shared across the $K$ optical-conditioning streams.
All streams interact through joint attention.
The subsequent single-stream blocks jointly process all tokens with shared parameters.
Only target tokens are passed to the final velocity head; the optical and SAR tokens provide conditioning information.

\paragraph{Conditional latent flow matching.}
For each target latent $\mathbf{z}^{\mathrm{ms}}$, we sample
an independent Gaussian source
$\boldsymbol{\epsilon}\sim\mathcal{N}(\mathbf{0},\mathbf{I})$ and a time $t\sim\mathcal{U}(0,1)$.
The linear path $\mathbf{z}_t$ and target velocity $\mathbf{u}$ are
\begin{equation}
    \mathbf{z}_t
    =(1-t)\boldsymbol{\epsilon}+t\mathbf{z}^{\mathrm{ms}},
    \qquad
    \mathbf{u}
    =\frac{\partial\mathbf{z}_t}{\partial t}
    =\mathbf{z}^{\mathrm{ms}}-\boldsymbol{\epsilon}.
    \label{eq:geocr_flow_path}
\end{equation}
The generator $\mathbf{v}_\theta$ predicts this velocity
conditioned on the available observations:
\begin{equation}
    \mathcal{L}_{\mathrm{FM}}
    =\mathbb{E}
    \left[
      \left\|
        \mathbf{v}_\theta(\mathbf{z}_t,t;\mathcal{C})
        -\mathbf{u}
      \right\|_2^2
    \right].
    \label{eq:geocr_flow_matching}
\end{equation}
Only the generator is optimized in this stage.
The expectation spans the heterogeneous training corpus and its
available observation configurations, so a common velocity
model $\mathbf{v}_\theta$ learns RGB and multispectral restoration from single-
and multi-temporal inputs with or without SAR guidance.

\subsection{Inference and Adaptation}
\label{sec:downstream_adaptation}

For each dataset, we use the shared Stage~2 generator while keeping all autoencoder trunks and stems frozen.
We consider two modes: (i) direct inference without dataset-specific optimization, denoted as \textbf{GeoCR (w/o FT)}, and (ii) low-rank adaptation (LoRA)~\citep{hu2021lora}, denoted as \textbf{GeoCR (LoRA)}.
Only the adapters are optimized on the target dataset's training
split using the Stage~2 flow-matching objective.
In both modes, inference integrates the learned conditional flow
from Gaussian noise and decodes the resulting RGB and, when
required, non-RGB latents through the corresponding frozen
decoder routes.

\begin{table*}[t]
    \scriptsize
    \centering
    \caption{\textbf{Overview of the GeoCR pretraining corpus.}
    Counts refer to cloud-free target images in the training splits.
    Spectral bands describe the targets unless otherwise noted;
    cloudy conditions indicate the number of optical input frames.
    Tile sizes are in pixels.
    L1C: Level-1C; TOA/BOA: top-/bottom-of-atmosphere reflectance;
    RGB: red, green, and blue; NIR: near-infrared;
    SAR: synthetic aperture radar; GSD: ground sampling distance.
    B8 and B10 denote Sentinel-2 band identifiers.}
    \vspace{-0.2cm}
    \label{tab:pretrain_corpus}
    \resizebox{1.0\textwidth}{!}{%
    \renewcommand{\arraystretch}{1.1}
    \newsavebox{\deliveredtilebox}
    \sbox{\deliveredtilebox}{%
        \makecell{\textbf{Delivered}\\\textbf{tile}}%
    }
    \setlength{\tabcolsep}{3pt}
    \begin{tabular}{lllccccr}
        \toprule
        \rowcolor{blue!5}
        \textbf{Source}
        & \textbf{Imagery source}
        & \textbf{Spectral bands}
        & \usebox{\deliveredtilebox}
        & \makecell{\textbf{GSD}\\\textbf{(m)}}
        & \makecell{\textbf{Cloudy}\\\textbf{cond.}}
        & \makecell{\textbf{SAR}\\\textbf{cond.}}
        & \makecell{\textbf{Clean}\\\textbf{targets}} \\
        \midrule
        AllClear~\citep{zhou2024allclear}
        & Sentinel-2 + Sentinel-1
        & 13 (L1C, TOA)
        & $256^2$ & 10 & 1--3 & \cmark & 662,022 \\

        SEN12MS-CR~\citep{ebel2021multisensor}
        & Sentinel-2 + Sentinel-1
        & 13 (L1C, TOA)
        & $256^2$ & 10 & 1 & \cmark & 107,143 \\

        WHUS2-CRv~\citep{li2022thin}
        & Sentinel-2
        & 13 (BOA; B10: L1C, TOA)
        & $384^2/192^2/64^2$ & 10 & 1 & \xmark & 18,816 \\

        Sen2\_MTC\_Old~\citep{sarukkai2020cloud}
        & Sentinel-2
        & 3 (RGB; +NIR for cloudy inputs)
        & $256^2$ & 10 & 3 & \xmark & 88,874 \\

        Sen2\_MTC\_New~\citep{huang2022ctgan}
        & Sentinel-2
        & 4 (RGB+B8, BOA)
        & $256^2$ & 10 & 3 & \xmark & 2,380 \\

        T-CLOUD~\citep{ding2022uncertainty}
        & Landsat-8
        & 3 (RGB, 8-bit)
        & $256^2$ & 30 & 1 & \xmark & 2,234 \\

        RICE2~\citep{lin2019rice}
        & Landsat-8
        & 3 (RGB, 8-bit)
        & $512^2$ & 30 & 1 & \xmark & 553 \\

        CUHK-CR1~\citep{sui2024diffusion}
        & Jilin-1 KF01B
        & 4 (RGB+NIR, 8-bit)
        & $512^2$ & 0.5 & 1 & \xmark & 508 \\

        CUHK-CR2~\citep{sui2024diffusion}
        & Jilin-1 KF01B
        & 4 (RGB+NIR, 8-bit)
        & $512^2$ & 0.5 & 1 & \xmark & 426 \\

        RICE1~\citep{lin2019rice}
        & Google Earth
        & 3 (RGB, 8-bit)
        & $512^2$ & -- & 1 & \xmark & 375 \\
        \bottomrule
    \end{tabular}}
\end{table*}

\begin{table}[tbp]
    \scriptsize
    \centering
    \caption{\textbf{Quantitative comparison on CUHK-CR2.}
    Its RGB evaluation setting supports both RGB-only image
    translation/restoration methods and specialized cloud removal
    methods, enabling comparison across all ten baselines.
    \textbf{Bold} and \underline{underlined} values indicate the
    best and second-best results, respectively.
    Additional results are provided in the \textit{Appendix}.}
    \vspace{-0.2cm}
    \label{tab:main_cuhk_cr2}
    \setlength{\tabcolsep}{4pt}
    \renewcommand{\arraystretch}{1.1}
    \resizebox{0.8\textwidth}{!}{%
    \begin{tabular}{llccccccc}
        \toprule
        \textbf{Method} & \textbf{Venue}
        & \textbf{FID}$\downarrow$ & \textbf{DISTS}$\downarrow$
        & \textbf{KID}$\downarrow$ & \textbf{DINO}$\uparrow$
        & \textbf{LPIPS}$\downarrow$ & \textbf{SSIM}$\uparrow$
        & \textbf{PSNR}$\uparrow$ \\
        \midrule
        \rowcolor{black!5}
        \multicolumn{9}{l}{\textit{General image translation/restoration methods}} \\
        pix2pix~\citep{isola2017image} & CVPR'17
        & 194.8 & 0.241 & 0.1201 & 0.538 & 0.319 & 0.452 & 19.46 \\
        pix2pixHD~\citep{wang2018high} & CVPR'18
        & 257.1 & 0.262 & 0.2130 & 0.379 & 0.304 & 0.401 & 19.34 \\
        BBDM~\citep{li2023bbdm} & CVPR'23
        & 514.6 & 0.542 & 0.5827 & 0.131 & 0.737 & 0.326 & 17.68 \\
        HI-Diff~\citep{chen2023hierarchical} & NeurIPS'23
        & 165.7 & 0.192 & 0.0769 & 0.664 & 0.253 & 0.638 & 23.54 \\
        \midrule
        \rowcolor{black!5}
        \multicolumn{9}{l}{\textit{Cloud removal methods}} \\
        UnCRtainTS~\citep{ebel2023uncrtaints} & CVPRW'23
        & 165.6 & 0.265 & 0.0761 & 0.529 & 0.345 & 0.586 & 22.12 \\
        DiffCR~\citep{zou2024diffcr} & TGRS'24
        & 245.2 & 0.285 & 0.1874 & 0.615 & 0.353 & 0.582 & 22.85 \\
        IDF-CR~\citep{wang2024idfcr} & TGRS'24
        & 167.0 & 0.214 & 0.0873 & 0.645 & 0.270 & \underline{0.641} & 23.18 \\
        ThiefCloud~\citep{zhao2025thiefcloud} & TCSVT'25
        & 136.3 & 0.226 & 0.0593 & 0.668 & 0.232 & 0.638 & \textbf{23.88} \\
        EMRDM~\citep{liu2025effective} & CVPR'25
        & 104.4 & 0.167 & 0.0259 & 0.727 & 0.208 & \textbf{0.654} & \underline{23.61} \\
        GACR~\citep{wang2026gacr} & ECCV'26
        & 125.0 & 0.178 & 0.0496 & 0.739 & 0.217 & 0.620 & 23.45 \\
        \midrule
        \rowcolor{blue!5}
        \multicolumn{9}{l}{\textit{Ours}} \\
        \textbf{GeoCR (w/o FT)} & --
        & \textbf{93.6} & \underline{0.153} & \underline{0.0200}
        & \textbf{0.784} & \textbf{0.194} & 0.607 & 23.21 \\
        \textbf{GeoCR (LoRA)} & --
        & \underline{94.7} & \textbf{0.152} & \textbf{0.0194}
        & \underline{0.779} & \underline{0.196} & 0.601 & 23.11 \\
        \bottomrule
    \end{tabular}}
\end{table}
\section{Experiments}
\label{sec:experiments}

\subsection{Pretraining and Evaluation Data}
\label{sec:data}

\paragraph{Multi-source pretraining corpus.}
We construct a unified corpus from the official training splits of ten cloud removal datasets, comprising 883,331 cloud-free optical targets.
Table~\ref{tab:pretrain_corpus} summarizes their sensing platforms, spectral bands, radiometric representations, and observation availability.
The corpus combines Sentinel-2, Landsat-8, and high-resolution imagery, including RGB and multispectral targets, single- and multi-temporal cloudy inputs, and optional SAR guidance.
These configurations jointly contribute to the shared GeoCR prior through the common latent interface.

\paragraph{Evaluation benchmarks.}
We build on the released training and test partitions, and apply the split corrections and duplicate exclusions documented in \textit{Appendix}. Evaluation uses the resulting held-out test samples.
Our five primary benchmarks include SEN12MS-CR, Sen2\_MTC\_New, CUHK-CR2, WHUS2-CRv, and T-CLOUD.
Full-band experiments retain the dataset-specific optical band configuration, while RGB-only experiments use RGB optical inputs and targets.
Within each setting, all methods are evaluated on identical test samples and spatial extents.
GeoCR (w/o FT) uses one jointly pretrained checkpoint across datasets; GeoCR (LoRA) adapts it using only the corresponding training split.
Additional benchmark results, test-set sizes, and detailed input configurations are provided in the
\textit{Appendix}.

\begin{table*}[tbp]
    \scriptsize
    \centering
    \caption{\textbf{Quantitative comparison on Sen2\_MTC\_New.}
    (a) Full-band setting. (b) RGB setting.}
    \vspace{-0.2cm}
    \label{tab:main_sen2mtc_new}
    \setlength{\tabcolsep}{3pt}
    \renewcommand{\arraystretch}{1.1}
    \begin{minipage}[t]{0.49\textwidth}
        \centering
        {\textbf{(a) Full-band (RGB + NIR)}}\\[2pt]
        \resizebox{\linewidth}{!}{%
        \begin{tabular}{lccccccc}
            \toprule
            \textbf{Method}
            & \textbf{FID}$\downarrow$
            & \textbf{DISTS}$\downarrow$
            & \textbf{KID}$\downarrow$
            & \textbf{DINO}$\uparrow$
            & \textbf{LPIPS}$\downarrow$
            & \textbf{SSIM}$\uparrow$
            & \textbf{PSNR}$\uparrow$ \\
            \midrule
            UnCRtainTS
            & 113.4 & 0.258 & 0.0597 & 0.471 & 0.428 & 0.588 & 17.15 \\
            DiffCR
            & 95.9 & 0.297 & 0.0388 & 0.499 & 0.312 & 0.599 & 19.26 \\
            ThiefCloud
            & 143.6 & 0.333 & 0.0767 & 0.310 & 0.508 & 0.481 & 14.92 \\
            EMRDM
            & 96.6 & 0.218 & 0.0459 & 0.533 & 0.314 & 0.640 & 18.20 \\
            GACR
            & 87.4 & 0.223 & 0.0358 & 0.535 & 0.320 & 0.626 & 18.90 \\
            \midrule
            \rowcolor{blue!5}
            \textbf{GeoCR (w/o FT)}
            & \underline{52.9} & \underline{0.182}
            & \underline{0.0032} & \underline{0.642}
            & \underline{0.221} & \textbf{0.661} & \textbf{20.77} \\
            \rowcolor{blue!5}
            \textbf{GeoCR (LoRA)}
            & \textbf{52.7} & \textbf{0.181}
            & \textbf{0.0031} & \textbf{0.643}
            & \textbf{0.220} & \underline{0.660} & \underline{20.76} \\
            \bottomrule
        \end{tabular}}
    \end{minipage}
    \hfill
    \begin{minipage}[t]{0.49\textwidth}
        \centering
        {\textbf{(b) RGB-only}}\\[2pt]
        \resizebox{\linewidth}{!}{%
        \begin{tabular}{lccccccc}
            \toprule
            \textbf{Method}
            & \textbf{FID}$\downarrow$
            & \textbf{DISTS}$\downarrow$
            & \textbf{KID}$\downarrow$
            & \textbf{DINO}$\uparrow$
            & \textbf{LPIPS}$\downarrow$
            & \textbf{SSIM}$\uparrow$
            & \textbf{PSNR}$\uparrow$ \\
            \midrule
            pix2pix
            & 161.6 & 0.305 & 0.1188 & 0.224 & 0.480 & 0.541 & 20.16 \\
            pix2pixHD
            & 153.9 & 0.292 & 0.1121 & 0.274 & 0.365 & 0.646 & 24.00 \\
            BBDM
            & 179.4 & 0.392 & 0.1243 & 0.243 & 0.504 & 0.652 & 24.48 \\
            HI-Diff
            & 125.8 & 0.302 & 0.0695 & 0.390 & 0.398
            & \textbf{0.742} & \textbf{25.66} \\
            \midrule
            \rowcolor{blue!5}
            \textbf{GeoCR (w/o FT)}
            & \underline{79.1} & \textbf{0.252}
            & \textbf{0.0223} & \underline{0.489}
            & \textbf{0.313} & 0.706 & \underline{25.41} \\
            \rowcolor{blue!5}
            \textbf{GeoCR (LoRA)}
            & \textbf{78.4} & \underline{0.265}
            & \underline{0.0313} & \textbf{0.520}
            & \underline{0.333} & \underline{0.712} & 25.07 \\
            \bottomrule
        \end{tabular}}
    \end{minipage}
\end{table*}

\begin{table*}[tbp]
    \scriptsize
    \centering
    \caption{\textbf{Quantitative comparison on SEN12MS-CR.}
    (a) Full-band setting. (b) RGB setting.}
    \vspace{-0.2cm}
    \label{tab:main_sen12mscr}
    \setlength{\tabcolsep}{3pt}
    \renewcommand{\arraystretch}{1.1}
    \begin{minipage}[t]{0.49\textwidth}
        \centering
        {\textbf{(a) Full-band (13 spectral bands)}}\\[2pt]
        \resizebox{\linewidth}{!}{%
        \begin{tabular}{lccccccc}
            \toprule
            \textbf{Method}
            & \textbf{FID}$\downarrow$
            & \textbf{DISTS}$\downarrow$
            & \textbf{KID}$\downarrow$
            & \textbf{DINO}$\uparrow$
            & \textbf{LPIPS}$\downarrow$
            & \textbf{SSIM}$\uparrow$
            & \textbf{PSNR}$\uparrow$ \\
            \midrule
            UnCRtainTS
            & 80.5 & 0.276 & 0.0497 & 0.440 & 0.313
            & \textbf{0.884} & 28.77 \\
            EMRDM
            & 75.7 & 0.274 & 0.0475 & 0.467 & 0.304
            & \underline{0.873} & 28.70 \\
            GACR
            & 96.9 & 0.331 & 0.0679 & 0.358 & 0.333 & 0.828 & 27.76 \\
            \rowcolor{blue!5}
            \textbf{GeoCR (w/o FT)}
            & \textbf{28.4} & \textbf{0.202}
            & \textbf{0.0091} & \textbf{0.571}
            & \textbf{0.215} & 0.866 & \textbf{29.29} \\
            \rowcolor{blue!5}
            \textbf{GeoCR (LoRA)}
            & \underline{29.5} & \underline{0.205}
            & \underline{0.0098} & \underline{0.569}
            & \underline{0.216} & 0.866 & \underline{29.19} \\
            \bottomrule
        \end{tabular}}
    \end{minipage}
    \hfill
    \begin{minipage}[t]{0.49\textwidth}
        \centering
        {\textbf{(b) RGB-only}}\\[2pt]
        \resizebox{\linewidth}{!}{%
        \begin{tabular}{lccccccc}
            \toprule
            \textbf{Method}
            & \textbf{FID}$\downarrow$
            & \textbf{DISTS}$\downarrow$
            & \textbf{KID}$\downarrow$
            & \textbf{DINO}$\uparrow$
            & \textbf{LPIPS}$\downarrow$
            & \textbf{SSIM}$\uparrow$
            & \textbf{PSNR}$\uparrow$ \\
            \midrule
            pix2pix
            & 145.9 & 0.317 & 0.1000 & 0.281 & 0.465 & 0.632 & 21.64 \\
            pix2pixHD
            & 70.8 & \textbf{0.239} & 0.0443 & 0.420
            & \textbf{0.277} & 0.772 & \underline{27.23} \\
            BBDM
            & 96.1 & 0.284 & 0.0536 & 0.331 & 0.367 & 0.710 & 25.44 \\
            HI-Diff
            & 62.7 & 0.284 & \underline{0.0342} & 0.438
            & 0.320 & \textbf{0.837} & \textbf{28.28} \\
            \rowcolor{blue!5}
            \textbf{GeoCR (w/o FT)}
            & \textbf{43.9} & \underline{0.244}
            & \textbf{0.0246} & \textbf{0.532}
            & 0.326 & 0.712 & 22.15 \\
            \rowcolor{blue!5}
            \textbf{GeoCR (LoRA)}
            & \underline{52.1} & 0.263
            & 0.0375 & \underline{0.529}
            & \underline{0.308} & \underline{0.781} & 24.59 \\
            \bottomrule
        \end{tabular}}
    \end{minipage}
\end{table*}

\begin{figure*}[t]
  \centering
  \includegraphics[width=0.8\textwidth]{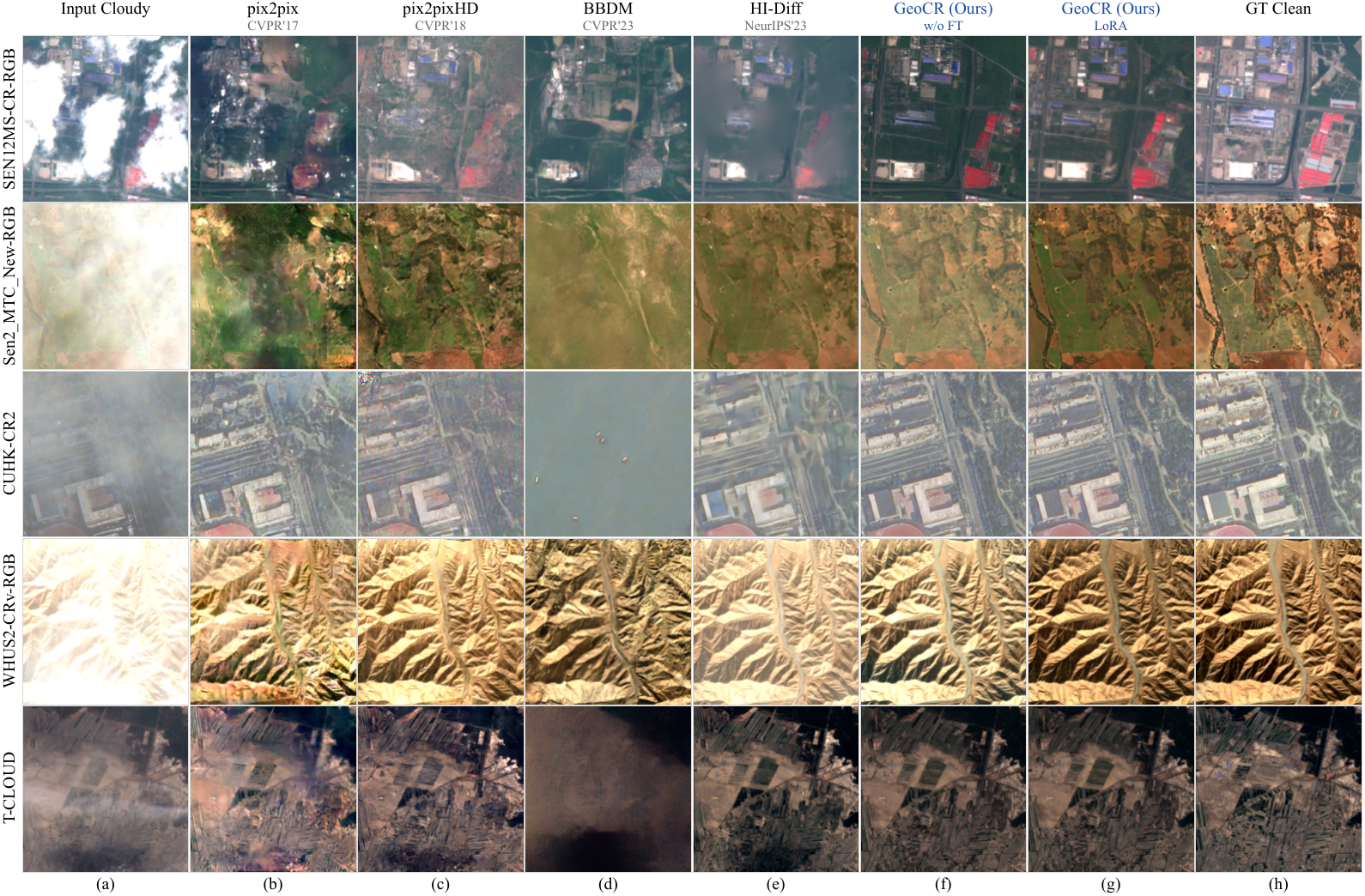}
  \vspace{-0.2cm}
  \caption{\textbf{Qualitative comparison with general restoration methods under RGB-only settings.}}
  \label{fig:result_v2}
\end{figure*}

\begin{figure*}[t]
    \centering
    \begin{minipage}[t]{0.29\textwidth}
        \vspace{0pt}
        \centering
        \includegraphics[width=\linewidth]{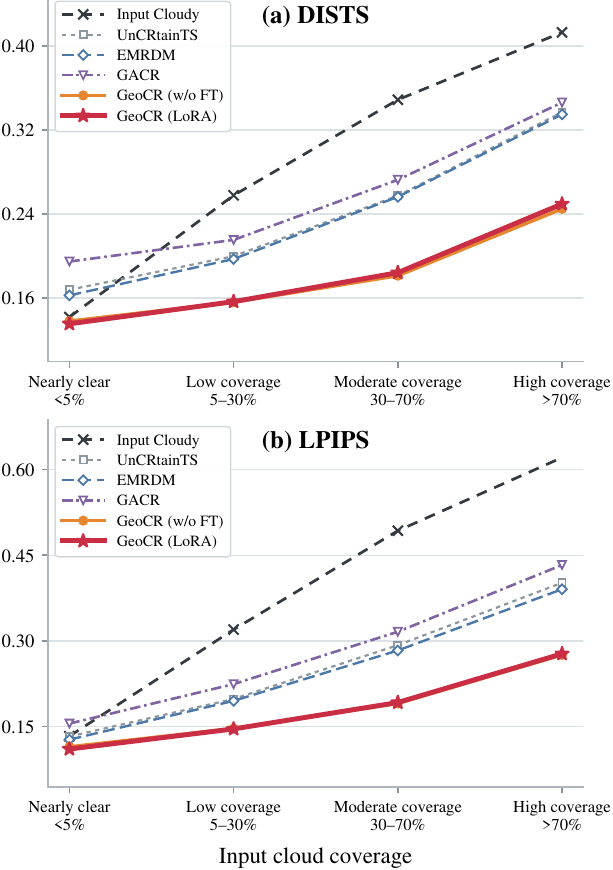}
    \end{minipage}
    \hfill
    \begin{minipage}[t]{0.67\textwidth}
        \vspace{0pt}
        \centering
        \includegraphics[width=\linewidth]{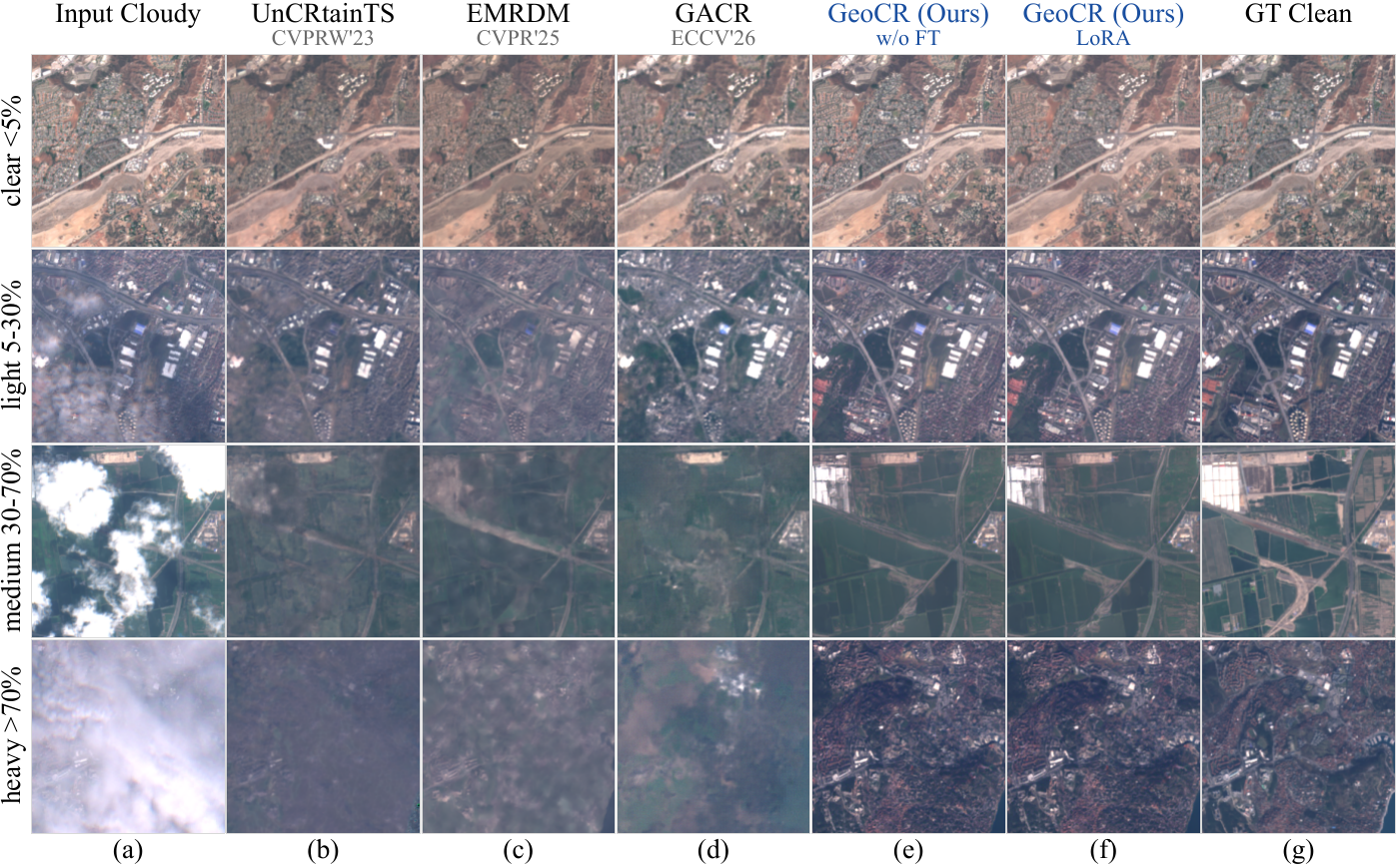}
    \end{minipage}
    \caption{\textbf{Cloud removal across input cloud coverage on SEN12MS-CR.}
    Left: DISTS and LPIPS grouped by input cloud coverage
    estimated with \texttt{s2cloudless}; lower is better.
    Input Cloudy denotes the unrestored observation.
    Right: qualitative comparisons across the same coverage groups.}
    \label{fig:cloud_coverage}
\end{figure*}

\subsection{Implementation Details and Evaluation Protocol}
\label{sec:implementation}

\paragraph{Implementation details.}
All models are trained on $256\times256$ crops and NVIDIA B200 GPUs.
Stage~1 trains 51.1K stem parameters around the frozen FLUX.2~\citep{flux-2-2025} VAE for 400K updates
(global batch size 64, two GPUs).
We use AdamW with a constant learning rate of $10^{-4}$,
without warmup or gradient clipping.
Stage~2 optimizes all 3.853B parameters of FLUX.2 4B
for 500K updates (global batch size 64, four GPUs) in bf16,
using AdamW with zero weight decay, a 2K-update linear warmup
to $10^{-4}$, and cosine decay to $10^{-6}$.
All conditions are replaced with a learned null token with
probability 0.1; SAR is independently dropped with probability
0.3, and non-RGB conditioning is dropped with probability 0.05.
Additionally, 5\% of AllClear batches use SAR-only conditioning.
Stages~1 and~2 take 47.0 and 195.3 hours, respectively.
Downstream LoRA starts from the Stage~2 checkpoint and uses
10K updates with global batch size 32 on one GPU.
With a learning rate of $10^{-4}$ and
$r_{\mathrm{L}}=\alpha_{\mathrm{L}}=16$, LoRA adapts all
80 transformer linear layers, updating 23.1M parameters (0.60\%).
For our GeoCR, inference uses only four Euler steps on a uniform time grid without classifier-free guidance and a fixed per-sample noise seed.

\paragraph{Baselines.}
We compare with ten methods spanning general image restoration
and specialized cloud removal.
General-purpose baselines comprise pix2pix~\citep{isola2017image},
pix2pixHD~\citep{wang2018high}, BBDM~\citep{li2023bbdm}, and
HI-Diff~\citep{chen2023hierarchical}.
Cloud removal baselines include
UnCRtainTS~\citep{ebel2023uncrtaints},
DiffCR~\citep{zou2024diffcr},
IDF-CR~\citep{wang2024idfcr},
ThiefCloud~\citep{zhao2025thiefcloud},
EMRDM~\citep{liu2025effective}, and
GACR~\citep{wang2026gacr}.
Baselines are retrained separately on each dataset's training
split using official implementations when available and assessed
under a common evaluation pipeline.
The reported baseline set follows the input compatibility of
each setting; CUHK-CR2's RGB setting supports all ten methods.
GeoCR's shared parent is pretrained on the pooled training splits.

\paragraph{Metrics.}
We use FID~\citep{heusel2017gans} and
DISTS~\citep{ding2020dists} as the primary measures of distributional
similarity and paired perceptual fidelity, respectively.
We additionally report KID~\citep{binkowski2018demystifying},
DINO feature similarity using a frozen DINOv3-SAT~\citep{simeoni2025dinov3},
LPIPS~\citep{zhang2018lpips}, SSIM~\citep{wang2004image}, and PSNR.
Together, these metrics assess generated-image distributions,
perceptual and feature-level correspondence, and reconstruction
fidelity to the paired cloud-free reference. All feature-based metrics are computed on RGB views.

\subsection{Comparison with the State of the Art}
\label{sec:main_results}

\paragraph{Quantitative comparison.}
Tables~\ref{tab:main_cuhk_cr2},~\ref{tab:main_sen2mtc_new}, and \ref{tab:main_sen12mscr} show that GeoCR achieves the best
FID and DISTS on full-band Sen2\_MTC\_New and SEN12MS-CR and RGB-only CUHK-CR2.
On full-band SEN12MS-CR, GeoCR (w/o FT) reduces FID by over 60\% relative to the strongest competing baseline;
GeoCR (LoRA) provides modest improvements on selected datasets.
Both GeoCR (w/o FT) and GeoCR (LoRA) also lead FID on RGB-only Sen2\_MTC\_New and SEN12MS-CR, although pix2pixHD retains lower DISTS and LPIPS on the latter.
We treat PSNR and SSIM as complementary measures: temporal appearance differences and cloud occlusion can make exact reference matching ambiguous, while minimizing squared pixel
error under uncertainty can favor oversmoothed
predictions~\citep{blau2018perception}.
We therefore emphasize FID and DISTS for distributional
quality and paired perceptual fidelity.

\paragraph{Qualitative comparison.}
Figures~\ref{fig:first} and~\ref{fig:result_v2} show clearer
terrain, field boundaries, and urban structures with less
residual cloud contamination.
These improvements are already evident in GeoCR (w/o FT),
supporting the shared prior's ability to restore local structure
across diverse observation settings.
Additional examples are provided in the \textit{Appendix}.

\subsection{Ablation Studies and Analysis}
\label{sec:diagnostics}

\paragraph{Performance across cloud coverage.}
Figure~\ref{fig:cloud_coverage} shows lower DISTS and LPIPS
for both GeoCR variants than the compared baselines across
SEN12MS-CR coverage groups defined solely from cloudy inputs.
In nearly clear scenes, GeoCR improves on the unrestored input,
while all compared baselines worsen DISTS.
Its advantage persists under heavy cloud cover, indicating
robust perceptual restoration across the evaluated coverage levels.

\begin{wraptable}{r}{0.62\textwidth}
    \vspace{-0.4cm}
    \scriptsize
    \centering
    \caption{\textbf{Effect of Stage~1 stem training.} Reconstruction PSNR (dB) on AllClear validation samples before and after stem training, with the pretrained encoder and decoder frozen. VV and VH denote SAR polarizations.}
    \vspace{-0.2cm}
    \label{tab:stem_gain}
    \setlength{\tabcolsep}{3pt}
    \renewcommand{\arraystretch}{1.1}
    \resizebox{\linewidth}{!}{%
    \begin{tabular}{lccccccccc}
        \toprule
        \multirow{2}{*}{\makecell{\textbf{Recon.}\\\textbf{PSNR}}}
        & \multicolumn{3}{c}{\textbf{RGB} (w/o training)}
        & \multicolumn{3}{c}{\textbf{Non-RGB} (stem training)}
        & \multicolumn{3}{c}{\textbf{SAR} (stem training)} \\
        \cmidrule(lr){2-4}
        \cmidrule(lr){5-7}
        \cmidrule(lr){8-10}
        & \textbf{All} & \textbf{Clear} & \textbf{Cloudy}
        & \textbf{All} & \textbf{Clear} & \textbf{Cloudy}
        & \textbf{All} & \textbf{VV} & \textbf{VH} \\
        \midrule
        Before
        & 45.78 & 46.77 & 45.11
        & 18.84 & 20.60 & 18.20
        & 18.40 & 18.57 & 18.24 \\
        After
        & 45.78 & 46.77 & 45.11
        & 40.08 & 41.41 & 38.34
        & 29.14 & 28.54 & 30.21 \\
        \midrule
        \rowcolor{blue!5}
        Gain
        & +0.00 & +0.00 & +0.00
        & \textbf{+21.24} & \textbf{+20.81} & \textbf{+20.14}
        & \textbf{+10.74} & \textbf{+9.97} & \textbf{+11.98} \\
        \bottomrule
    \end{tabular}}
    \vspace{-0.2cm}
\end{wraptable}

\paragraph{Effect of stem training.}
With the pretrained trunks frozen, stem training improves
reconstruction PSNR by 21.24\,dB for ten-band non-RGB inputs
and 10.74\,dB for SAR (Table~\ref{tab:stem_gain}).
Gains across clear and cloudy observations and both SAR polarizations
support adapting lightweight interfaces to different measurement
domains.
The RGB path already provides high reconstruction fidelity
without adaptation.
Similarly, the initialized single-band NIR stem achieves
42.06\,dB on Sen2\_MTC\_New, 40.41\,dB on CUHK-CR, and
49.51\,dB on Sen2\_MTC\_Old without stem training,
supporting its reuse without further optimization.

\begin{figure*}[t]
  \centering
  \begin{minipage}{0.48\textwidth}
  \centering
  \includegraphics[width=\textwidth]{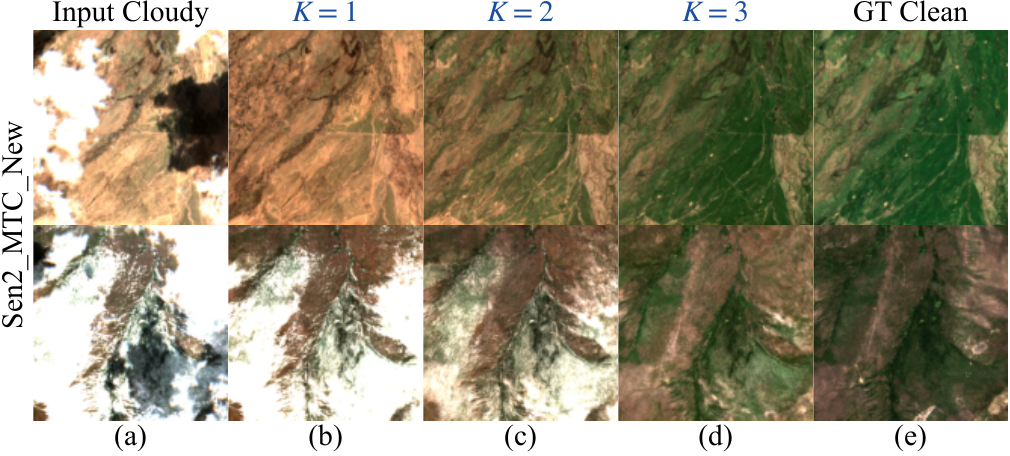}
  \vspace{-0.6cm}
  \caption{\textbf{Temporal conditioning on Sen2\_MTC\_New.}
  The same GeoCR (w/o FT) checkpoint uses $K$ cloudy
  optical frames.}
  \label{fig:ablation_cloudy}
  \end{minipage}
  \hfill
  \begin{minipage}{0.48\textwidth}
  \centering
  \includegraphics[width=\textwidth]{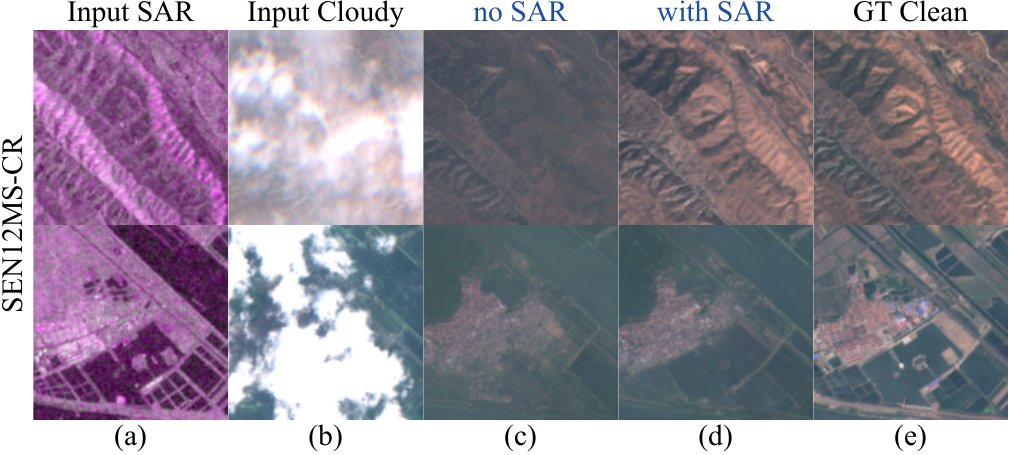}
  \vspace{-0.6cm}
  \caption{\textbf{SAR conditioning on SEN12MS-CR.}
  GeoCR (w/o FT) uses identical cloudy optical inputs
  with or without SAR guidance.}
  \label{fig:ablation_sar}
  \end{minipage}
\end{figure*}

\begin{wraptable}{r}{0.63\textwidth}
    \vspace{-0.4cm}
    \scriptsize
    \centering
    \begin{minipage}[t]{0.48\linewidth}
        \centering
        \caption{\textbf{Temporal conditioning on Sen2\_MTC\_New.}
        GeoCR uses the shared checkpoint without fine-tuning.}
        \vspace{-0.2cm}
        \label{tab:ablation_temporal}
        \setlength{\tabcolsep}{3pt}
        \renewcommand{\arraystretch}{1.1}
        \resizebox{0.8\linewidth}{!}{%
        \begin{tabular}{lccc}
            \toprule
            \textbf{Method}
            & \textbf{Cloudy}
            & \textbf{PSNR}$\uparrow$
            & \textbf{SSIM}$\uparrow$ \\
            \midrule
            GeoCR
            & 1 & 16.390 & 0.4595 \\
            GeoCR
            & 2 & 19.319 & 0.5963 \\
            \rowcolor{blue!5}
            GeoCR
            & 3 & \textbf{20.522} & \textbf{0.6568} \\
            \bottomrule
        \end{tabular}}
    \end{minipage}
    \hfill
    \begin{minipage}[t]{0.48\linewidth}
        \centering
        \caption{\textbf{SAR conditioning on SEN12MS-CR.}
        GeoCR uses the shared checkpoint without fine-tuning.}
        \vspace{-0.2cm}
        \label{tab:ablation_sar}
        \setlength{\tabcolsep}{3pt}
        \renewcommand{\arraystretch}{1.1}
        \resizebox{0.8\linewidth}{!}{%
        \begin{tabular}{lccc}
            \toprule
            \textbf{Method}
            & \textbf{SAR}
            & \textbf{PSNR}$\uparrow$
            & \textbf{SSIM}$\uparrow$ \\
            \midrule
            GeoCR
            & \xmark & 28.596 & 0.8572 \\
            \rowcolor{blue!5}
            GeoCR
            & \cmark & \textbf{29.442} & \textbf{0.8703} \\
            \bottomrule
        \end{tabular}}
    \end{minipage}
    \vspace{-0.2cm}
\end{wraptable}

\paragraph{Effect of temporal and SAR conditioning.}
Tables~\ref{tab:ablation_temporal} and~\ref{tab:ablation_sar}
vary the available observations using the same GeoCR
(w/o FT) checkpoint on fixed test subsets.
Additional cloudy frames and SAR guidance improve reference
fidelity, accompanied by reduced residual clouds and clearer
terrain ridges, respectively
(Figures~\ref{fig:ablation_cloudy} and~\ref{fig:ablation_sar}).
These results support using complementary evidence within
one model without retraining for each input configuration.
\section{Conclusion}
We presented GeoCR, a generalist model that learns a shared cloud removal prior through joint pretraining on ten datasets comprising 883,331 cloud-free target images.
Its shared latent interface enables a single generator to accommodate diverse spectral bands, temporal observations, and optional SAR guidance.
GeoCR achieves strong distributional and perceptual quality without dataset-specific fine-tuning and supports lightweight LoRA adaptation, by \textit{significantly} outperforming existing CR methods.
Conditioning ablations show that the same model benefits from additional temporal and SAR evidence without retraining.
These results demonstrate the feasibility of learning a reusable cloud removal model across the evaluated heterogeneous observation settings.

\clearpage
\providecommand{\geow}{GeoCR (w/o FT)}
\providecommand{\geol}{GeoCR (LoRA)}

\appendix
\section*{Appendix}
This \textit{Appendix} provides further analysis of
GeoCR's shared latent interface, implementation details, additional
results, and evaluation protocols. Table~\ref{tab:supple_overview}
summarizes its organization.

\begin{table}[h]
    \scriptsize
    \centering
    \caption{Overview of the \textit{Appendix}.}
    \label{tab:supple_overview}
    \resizebox{0.45\linewidth}{!}{%
    \renewcommand{\arraystretch}{1.1}
    \setlength{\tabcolsep}{6pt}
    \begin{tabular}{cl}
        \toprule
        \rowcolor{blue!5}
        \textbf{Section} & \textbf{Contents} \\
        \midrule
        Section~\ref{sec:supp_stem} & Stem adaptation \\
        Section~\ref{app:implementation} & Implementation details \\
        Section~\ref{sec:supp_quant} & Additional quantitative results \\
        Section~\ref{sec:supp_qual} & Additional qualitative results \\
        Section~\ref{sec:supp_limits} & Failure cases and limitations \\
        Section~\ref{sec:supp_data} & Datasets and preprocessing \\
        Section~\ref{sec:supp_eval} & Evaluation protocols \\
        \bottomrule
    \end{tabular}}
\end{table}

\section{Stem Adaptation}
\label{sec:supp_stem}

\begin{figure}[tbp]
    \centering
    \includegraphics[width=\linewidth]{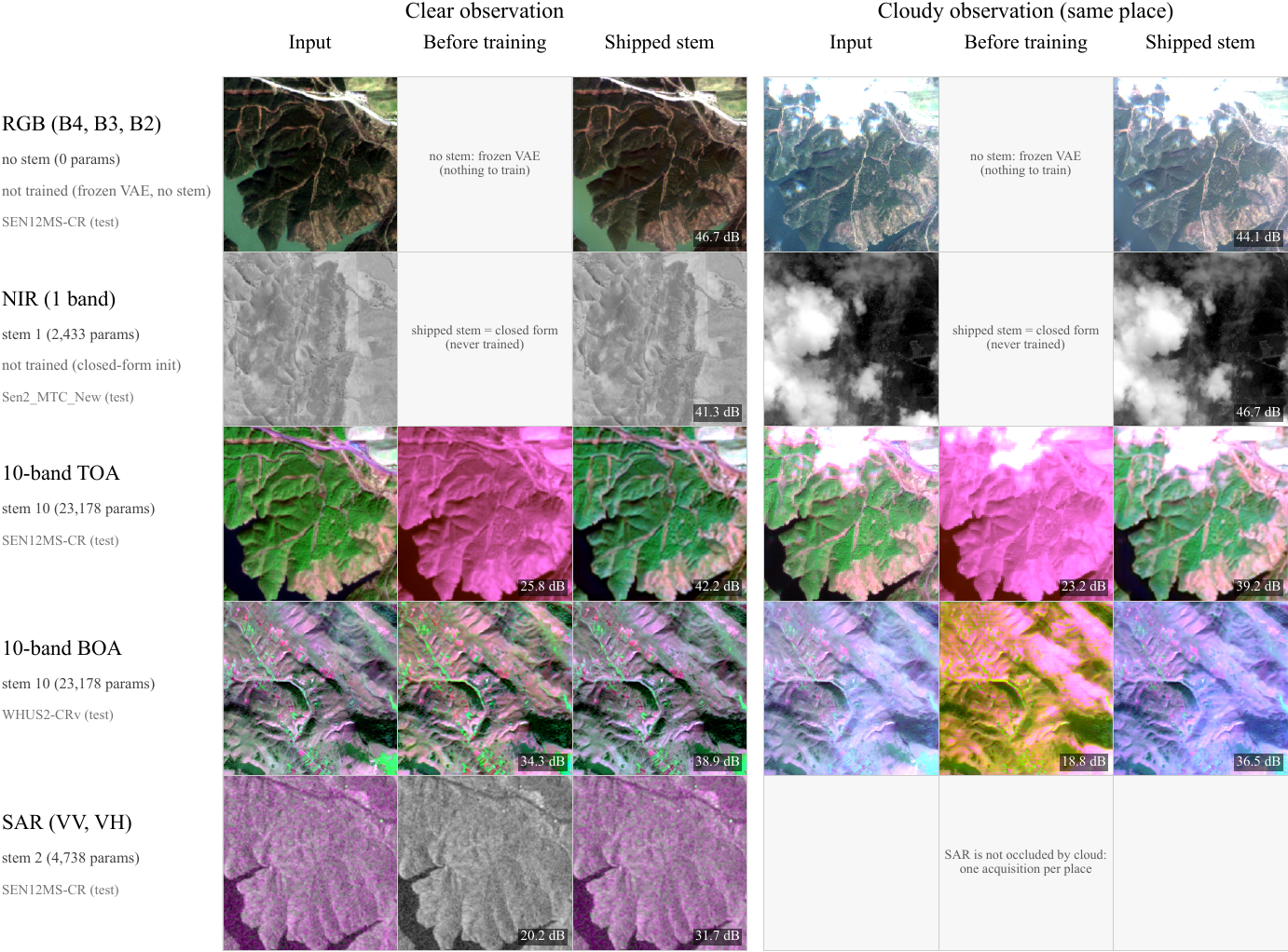}
    \caption{\textbf{Reconstruction through the shared autoencoder.}
    The pretrained RGB route and initialized NIR route
    remain fixed during Stage~1. Ten-band optical and SAR
    routes use learned stems with frozen encoder--decoder trunks.
    Ten-band visualizations use B12/B8/B5 false color, and SAR uses a VV/VH/VV
    composite.}
    \label{fig:supp_stems}
\end{figure}

\paragraph{Reusing the pretrained representation.}
Stage~1 adapts compact input and output stems while keeping the
pretrained RGB encoder--decoder trunks frozen.
Figure~\ref{fig:supp_stems} compares reconstruction across RGB,
single-band NIR, ten-band optical, and dual-polarization SAR inputs.
The trained stems improve reconstruction of measurements whose
channel layouts and statistics differ from RGB, while the original
RGB route remains unchanged.
The single-band NIR route retains its initialization and requires
no stem training.

\paragraph{Reconstruction across configurations.}
Table~\ref{tab:supp_routes} complements the held-out stem-training
ablation in the main paper with per-source reconstruction diagnostics.
The initialized NIR route achieves 42.06\,dB on Sen2\_MTC\_New,
40.41\,dB on CUHK-CR, and 49.51\,dB on Sen2\_MTC\_Old.
These diagnostics support retaining the initialized NIR interface
while adapting stems for ten-band optical and SAR observations.
They measure reconstruction of each input domain, rather than
cloud removal or preservation of every spectral relationship.

\section{Implementation Details}
\label{app:implementation}
\label{sec:supp_impl}

\paragraph{Autoencoder.}
All observation routes share the frozen encoder and decoder trunks
of the pretrained FLUX.2 RGB autoencoder.
Each input and output stem consists of a single convolutional layer.
We retain the original RGB projections and the initialized
single-band NIR stems, and train separate stems for ten-band TOA,
ten-band BOA, and VV/VH SAR inputs using an $\ell_1$
reconstruction loss.
The added stems contain 53.5K parameters, of which 51.1K are trained.
The encoder produces 32-channel posterior-mean latents at $1/8$
spatial resolution, which are packed into 128 channels at $1/16$
resolution and normalized using fixed BatchNorm statistics.
Decoding reverses normalization and packing.
All stems and trunks remain frozen after Stage~1.

\paragraph{Generator.}
GeoCR builds on FLUX.2 [klein] 4B Base~\citep{flux-2-2025},
with five double-stream and twenty single-stream blocks,
a hidden dimension of 3,072, and 24 attention heads.
RGB and available non-RGB latents are concatenated along channels
within each optical observation.
The noisy target, each of the $K$ cloudy observations, and optional
SAR observations retain separate token sequences, yielding
$1+K+\mathbf{1}_{\mathrm{SAR}}$ token streams.
The cloudy frames are therefore combined through attention rather
than channel-wise concatenation across time.
Double-stream blocks allow the target and conditioning tokens to
interact through joint attention; subsequent single-stream blocks
process their combined sequence with shared parameters.
Only target tokens are passed to the final velocity head.
GeoCR (w/o FT) uses the shared pretrained checkpoint, whereas
GeoCR (LoRA) learns dataset-specific low-rank updates~\citep{hu2021lora} while
keeping the autoencoder fixed.

\section{Additional Quantitative Results}
\label{sec:supp_quant}
Table~\ref{tab:supp_additional} extends the main comparisons to
T-CLOUD, CUHK-CR1, and WHUS2-CRv.
The native and RGB-only settings follow the preprocessing and
metric definitions in Sections~\ref{sec:supp_data}
and~\ref{sec:supp_eval}.
Bold and underlined values indicate the best and second-best
results.

\begin{table}[tbp]
\scriptsize\centering
\caption{\textbf{Reconstruction by configuration.} PSNR (dB) is computed from pooled reconstruction error on 64 training images per source, without clipping. These diagnostics differ from the held-out before/after comparison in the main paper and do not measure cloud removal.}
\label{tab:supp_routes}
\renewcommand{\arraystretch}{1.1}\setlength{\tabcolsep}{4pt}
\resizebox{\linewidth}{!}{\begin{tabular}{llcl}
\toprule\rowcolor{blue!5}
\textbf{Route}&\textbf{Stem parameters}&\textbf{Trained}&\textbf{Reconstruction PSNR}\\
\midrule
10-band TOA&23,178&Yes&AllClear 38.23; SEN12MS-CR 37.69\\
10-band BOA&23,178&Yes&WHUS2-CRv 37.89\\
Single-band NIR&2,433&No&Sen2\_MTC\_New 42.06; CUHK-CR 40.41; Sen2\_MTC\_Old 49.51\\
SAR VV/VH&4,738&Yes&AllClear 28.51; SEN12MS-CR 31.19\\
\bottomrule\end{tabular}}
\end{table}

\begin{table}[tbp]
    \scriptsize
    \centering
    \caption{\textbf{Additional quantitative comparisons.}
    Feature metrics use RGB views; PSNR/SSIM follow
    Table~\ref{tab:supp_eval}.}
    \label{tab:supp_additional}
    \setlength{\tabcolsep}{5pt}
    \renewcommand{\arraystretch}{1.0}
    \resizebox{0.75\linewidth}{!}{%
    \begin{tabular}{lccccccc}
        \toprule
        \textbf{Method}
        & \textbf{FID}$\downarrow$
        & \textbf{DISTS}$\downarrow$
        & \textbf{KID}$\downarrow$
        & \textbf{DINO}$\uparrow$
        & \textbf{LPIPS}$\downarrow$
        & \textbf{SSIM}$\uparrow$
        & \textbf{PSNR}$\uparrow$ \\
        \midrule
        \rowcolor{black!5}
        \multicolumn{8}{l}{\textbf{(a) T-CLOUD: RGB display imagery}} \\
        pix2pix~\citep{isola2017image}
        & 107.6 & 0.254 & 0.0509 & 0.450 & 0.330 & 0.635 & 20.13 \\
        pix2pixHD~\citep{wang2018high}
        & 60.8 & 0.169 & 0.0173 & 0.616 & 0.177 & 0.741 & 24.75 \\
        BBDM~\citep{li2023bbdm}
        & 186.6 & 0.422 & 0.1179 & 0.259 & 0.546 & 0.541 & 22.69 \\
        HI-Diff~\citep{chen2023hierarchical}
        & 40.2 & 0.124 & 0.0070 & 0.726 & 0.114 & \textbf{0.874} & \textbf{30.54} \\
        UnCRtainTS~\citep{ebel2023uncrtaints}
        & 63.3 & 0.180 & 0.0188 & 0.640 & 0.178 & 0.809 & 26.28 \\
        DiffCR~\citep{zou2024diffcr}
        & 139.6 & 0.358 & 0.0732 & 0.376 & 0.482 & 0.402 & 20.67 \\
        IDF-CR~\citep{wang2024idfcr}
        & 84.4 & 0.220 & 0.0304 & 0.467 & 0.236 & 0.787 & 25.99 \\
        ThiefCloud~\citep{zhao2025thiefcloud}
        & 40.4 & 0.123 & 0.0058 & 0.726 & \textbf{0.109} & 0.861 & 29.36 \\
        EMRDM~\citep{liu2025effective}
        & \underline{36.5} & \underline{0.121} & 0.0045 & \underline{0.759}
        & \underline{0.110} & \underline{0.867} & 28.25 \\
        GACR~\citep{wang2026gacr}
        & 39.2 & \textbf{0.120} & 0.0058 & 0.747 & 0.113 & 0.858 & \underline{29.56} \\
        \rowcolor{blue!5}
        \geow
        & \underline{36.5} & 0.126 & \underline{0.0024} & \textbf{0.763}
        & 0.122 & 0.785 & 27.15 \\
        \rowcolor{blue!5}
        \geol
        & \textbf{36.3} & 0.125 & \textbf{0.0022} & 0.751
        & 0.122 & 0.786 & 27.23 \\
        \midrule
        \rowcolor{black!5}
        \multicolumn{8}{l}{\textbf{(b) CUHK-CR1: RGB+NIR setting}} \\
        UnCRtainTS
        & 134.8 & 0.204 & 0.0283 & 0.712 & 0.311 & 0.679 & 23.82 \\
        DiffCR
        & 237.6 & 0.291 & 0.1360 & 0.649 & 0.318 & 0.573 & 22.75 \\
        EMRDM
        & \textbf{77.8} & \textbf{0.115} & -0.0006 & \textbf{0.845}
        & \textbf{0.146} & \textbf{0.760} & \textbf{25.64} \\
        GACR
        & 97.1 & 0.133 & 0.0099 & \underline{0.825}
        & \underline{0.160} & \underline{0.726} & \underline{25.02} \\
        \rowcolor{blue!5}
        \geow
        & \underline{80.4} & \underline{0.125} & \underline{-0.0025}
        & \underline{0.825} & 0.165 & 0.680 & 23.88 \\
        \rowcolor{blue!5}
        \geol
        & 81.5 & \underline{0.125} & \textbf{-0.0027}
        & 0.821 & 0.167 & 0.677 & 23.83 \\
        \midrule
        \rowcolor{black!5}
        \multicolumn{8}{l}{\textbf{(c) WHUS2-CRv: native multispectral setting}} \\
        UnCRtainTS
        & 26.9 & 0.142 & 0.0062 & 0.808 & 0.139 & \underline{0.926} & 31.14 \\
        IDF-CR
        & 40.0 & 0.179 & 0.0115 & 0.691 & 0.188 & 0.858 & 29.00 \\
        EMRDM
        & \textbf{16.9} & \textbf{0.101} & \textbf{0.0009} & \textbf{0.875}
        & \textbf{0.097} & \textbf{0.937} & \textbf{32.55} \\
        GACR
        & 22.6 & 0.230 & 0.0033 & 0.789 & 0.138 & 0.882 & 31.09 \\
        \rowcolor{blue!5}
        \geow
        & 18.9 & 0.111 & 0.0027 & 0.850 & 0.106 & 0.905 & \underline{32.29} \\
        \rowcolor{blue!5}
        \geol
        & \underline{18.5} & \underline{0.107} & \underline{0.0025}
        & \underline{0.858} & \underline{0.103} & 0.903 & 32.15 \\
        \midrule
        \rowcolor{black!5}
        \multicolumn{8}{l}{\textbf{(d) WHUS2-CRv: RGB-only setting}} \\
        pix2pix
        & 95.1 & 0.238 & 0.0523 & 0.432 & 0.274 & 0.746 & 23.75 \\
        pix2pixHD
        & 27.9 & 0.148 & 0.0057 & 0.751 & 0.145 & 0.828 & 26.77 \\
        BBDM
        & 50.4 & 0.213 & 0.0171 & 0.413 & 0.291 & 0.639 & 25.19 \\
        HI-Diff
        & \textbf{17.8} & \textbf{0.107} & \textbf{0.0018} & \textbf{0.853}
        & \textbf{0.096} & \textbf{0.894} & \textbf{29.91} \\
        \rowcolor{blue!5}
        \geow
        & 21.6 & \underline{0.137} & \underline{0.0032} & 0.786
        & \underline{0.133} & \underline{0.833} & \underline{27.16} \\
        \rowcolor{blue!5}
        \geol
        & \underline{20.7} & 0.138 & 0.0039 & \underline{0.811}
        & 0.134 & 0.804 & 25.35 \\
        \bottomrule
    \end{tabular}}
\end{table}

\paragraph{Performance across observation settings.}
On T-CLOUD, specialized methods retain advantages in perceptual
or pixel-level fidelity.
On CUHK-CR1, GeoCR achieves competitive FID and the lowest KID,
while EMRDM performs better on DISTS, LPIPS, SSIM, and PSNR.
On WHUS2-CRv, GeoCR remains competitive in distributional and
perceptual quality, although EMRDM leads the native setting and
HI-Diff leads the RGB-only setting.
LoRA provides selective improvements rather than consistent gains
over direct inference.
Together, these results show the reach of a shared cloud removal
prior while identifying settings that remain challenging.

\section{Additional Qualitative Results}
\label{sec:supp_qual}

\begin{figure}[tbp]
    \centering
    \includegraphics[width=\linewidth]{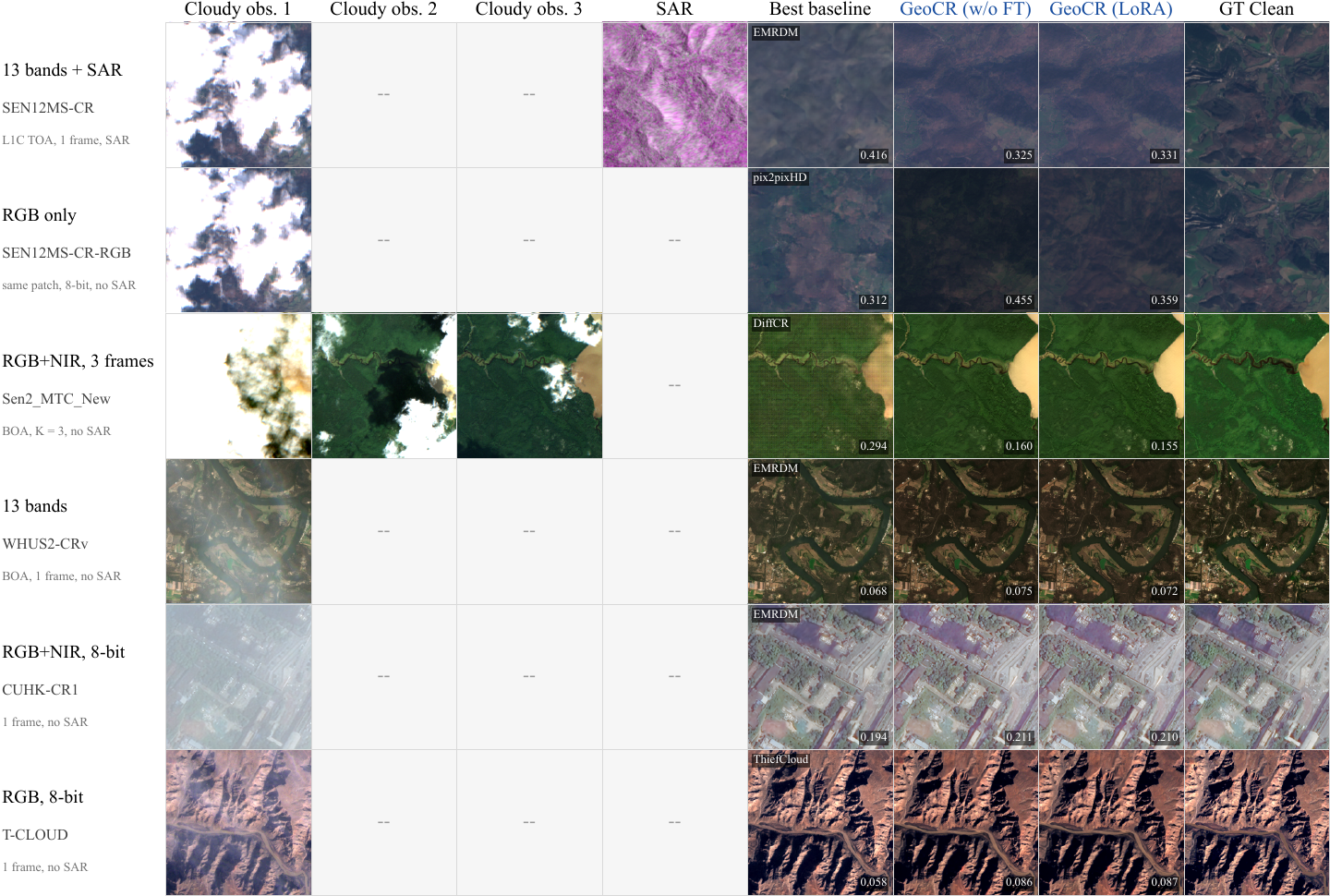}
    \caption{\textbf{Representative cloud removal results.}
    Rows show native and RGB-only SEN12MS-CR, Sen2\_MTC\_New,
    WHUS2-CRv, CUHK-CR1, and T-CLOUD.
    Baselines are selected by dataset-level LPIPS; numbers denote
    per-image LPIPS.
    The first two rows show the same scene.}
    \label{fig:supp_representative}
\end{figure}

\paragraph{Perceptual quality and reference fidelity.}
Figure~\ref{fig:supp_representative} compares the two GeoCR modes
with strong dataset-specific baselines across different observation
configurations.
The examples illustrate recovery of scene layout and local appearance,
as well as cases where a specialized baseline better matches the
reference. LoRA produces relatively small changes in these examples
and does not improve every image.
The two SEN12MS-CR rows differ in both available observations and
rendering, so their contrast does not isolate the contribution of SAR.

\begin{figure}[t]
    \centering
    \includegraphics[width=\linewidth]{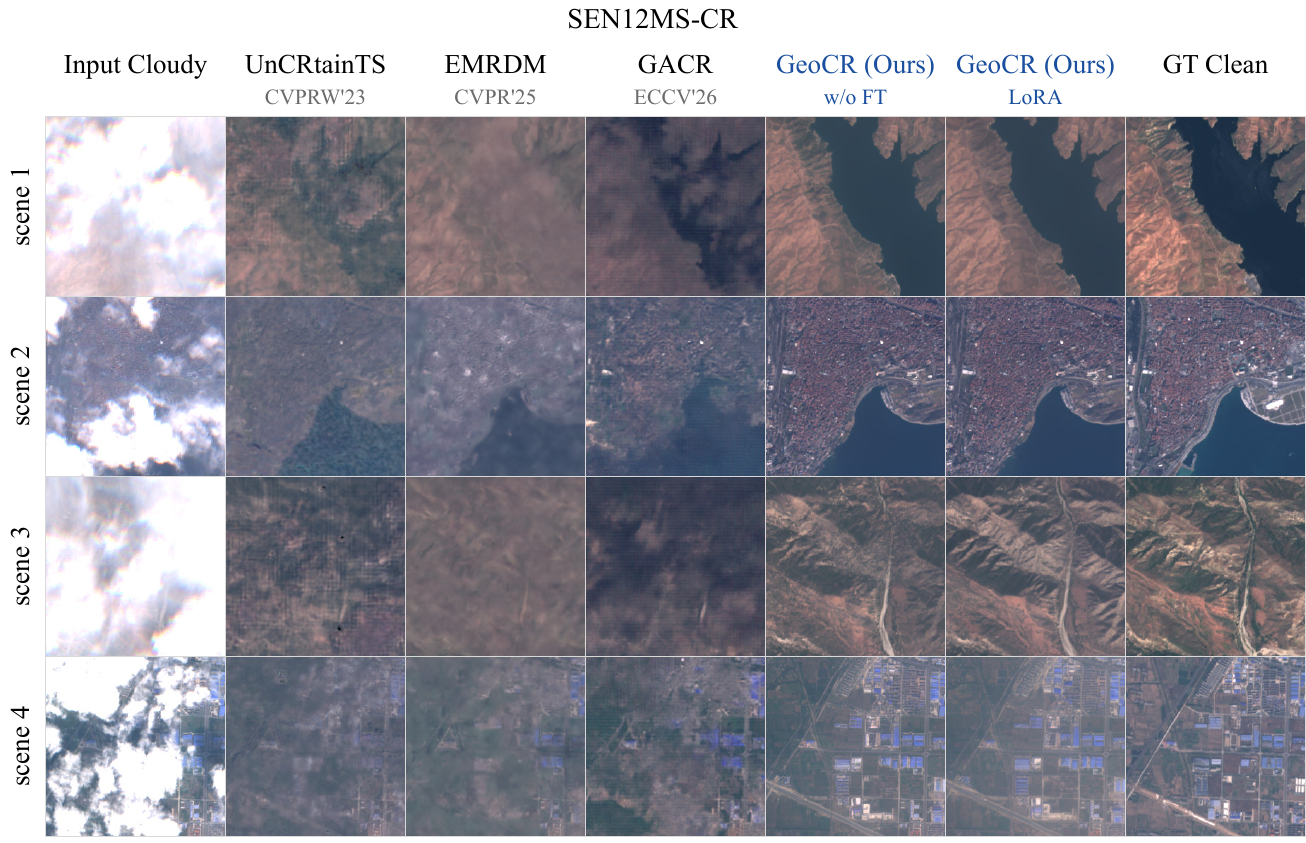}
    \caption{\textbf{Qualitative comparison on SEN12MS-CR~\citep{ebel2021multisensor}.} Four examples from the native multispectral setting, shown as RGB views with paired clear references.
    GeoCR is evaluated without fine-tuning (w/o FT) and with LoRA adaptation.}
    \label{fig:qual_sen12mscr}
\end{figure}

\begin{figure}[t]
    \centering
    \includegraphics[width=\linewidth]{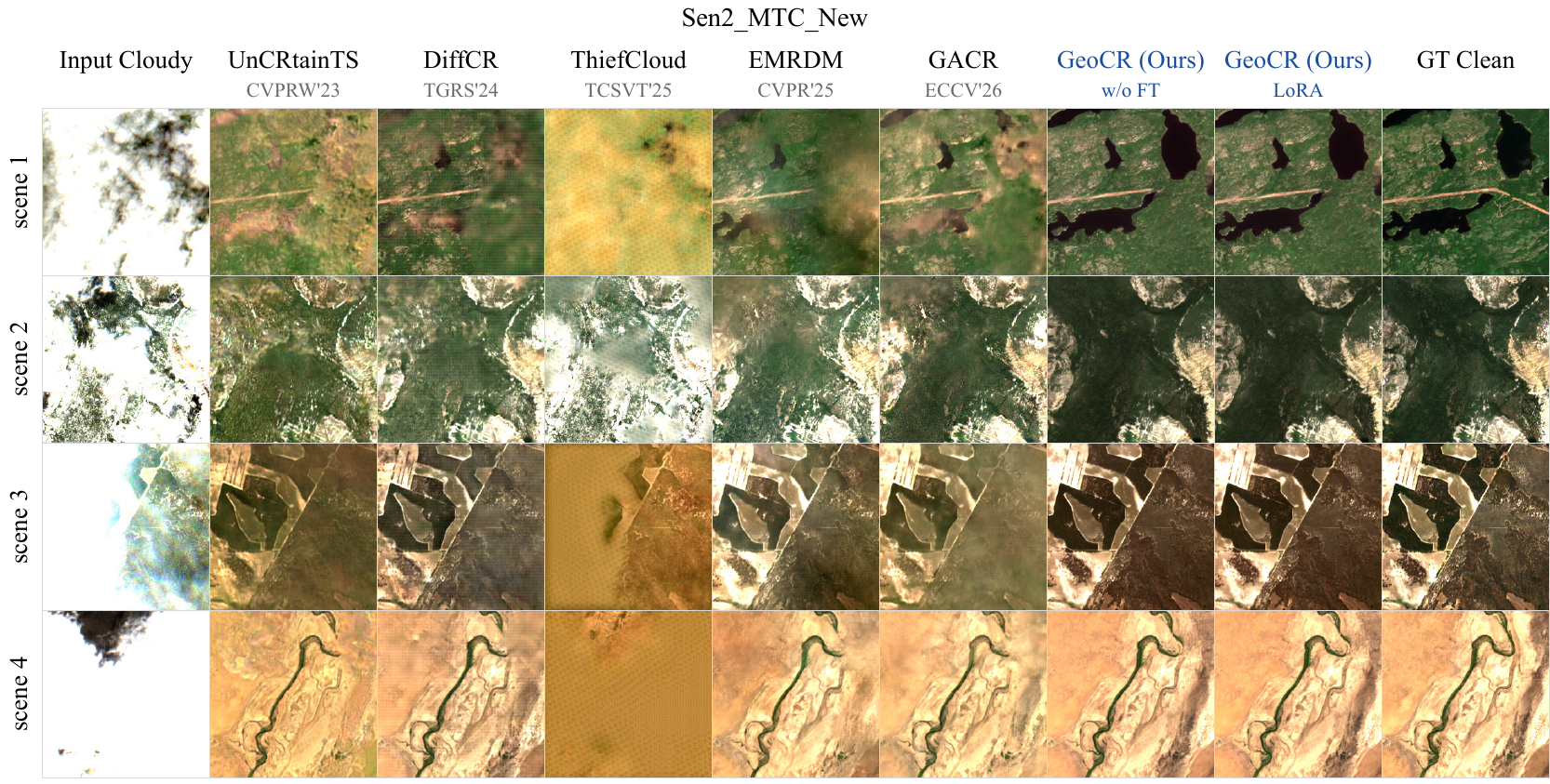}
    \caption{\textbf{Qualitative comparison on Sen2\_MTC\_New~\citep{huang2022ctgan}.} Four examples from the native multi-temporal setting, shown as RGB views with paired clear references.
    GeoCR is evaluated without fine-tuning (w/o FT) and with LoRA adaptation.}
    \label{fig:qual_mtc_new}
\end{figure}

\begin{figure}[t]
    \centering
    \includegraphics[width=\linewidth]{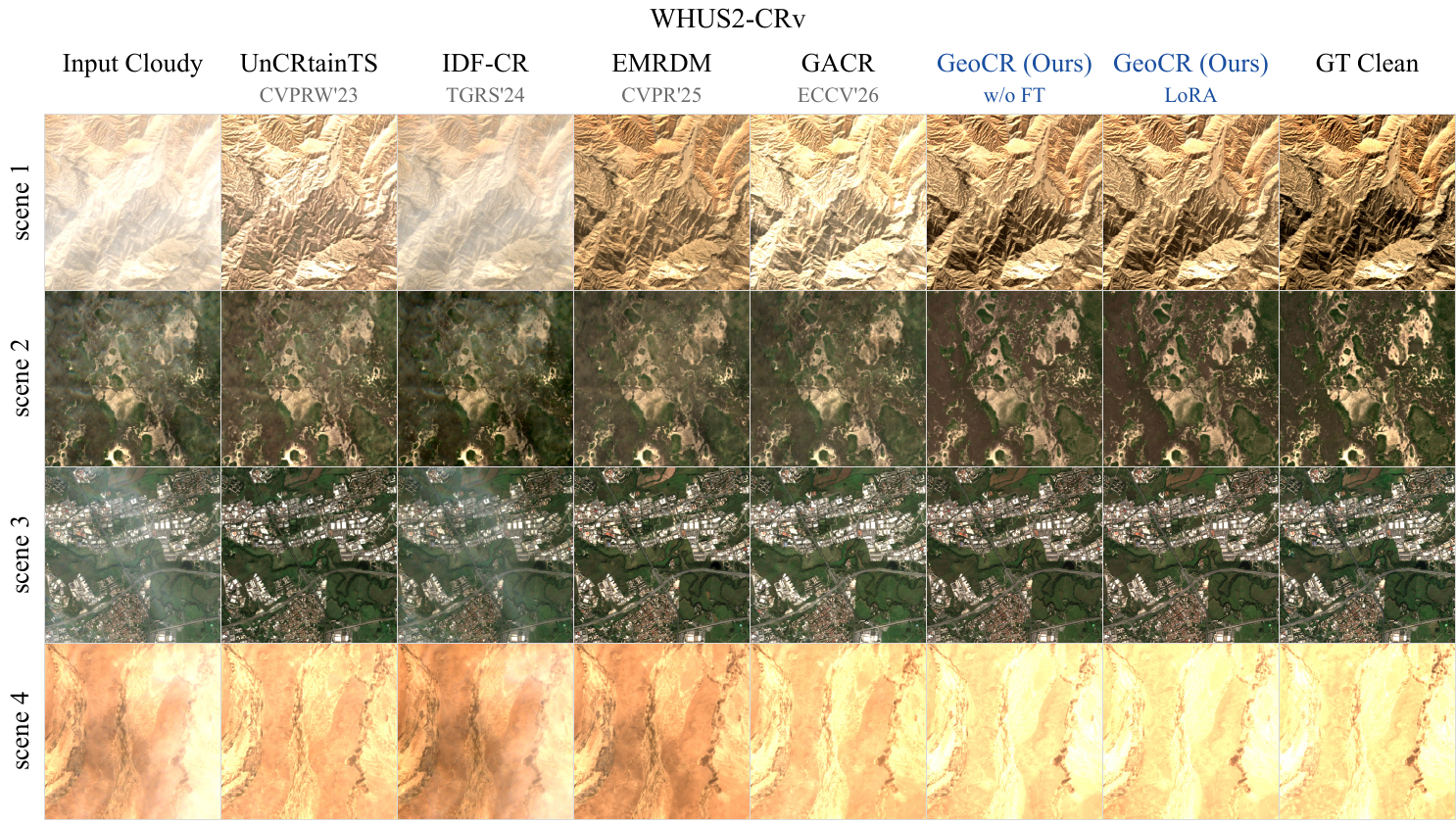}
    \caption{\textbf{Qualitative comparison on WHUS2-CRv~\citep{li2022thin}.} Four examples from the native multispectral setting, shown as RGB views with paired clear references.
    GeoCR is evaluated without fine-tuning (w/o FT) and with LoRA adaptation.}
    \label{fig:qual_whus}
\end{figure}

\begin{figure}[t]
    \centering
    \includegraphics[width=\linewidth]{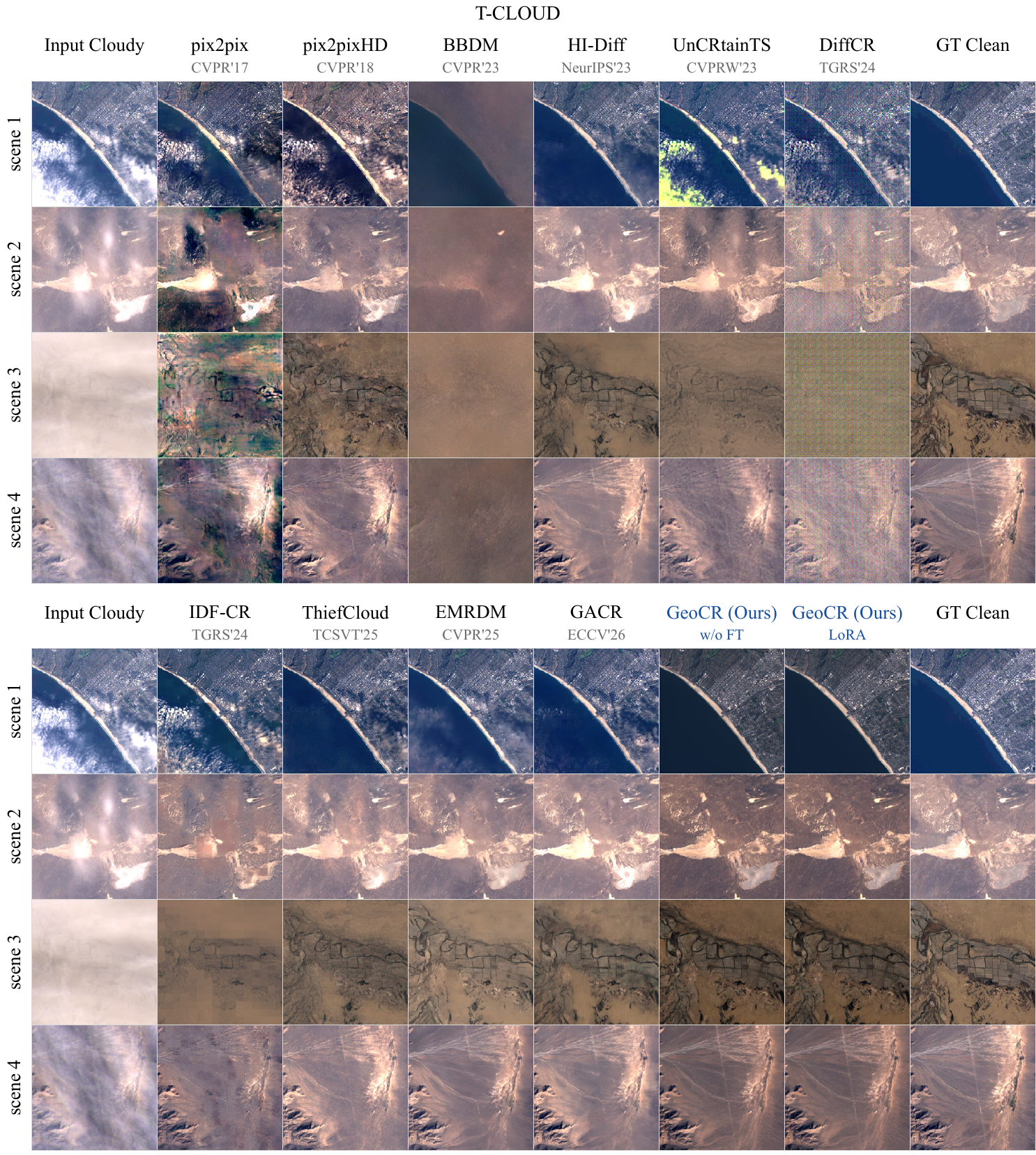}
    \caption{\textbf{Qualitative comparison on T-CLOUD~\citep{ding2022uncertainty}.} The two panels show the same four scenes with different groups of competing methods.
    GeoCR is evaluated without fine-tuning (w/o FT) and with LoRA adaptation.}
    \label{fig:qual_tcloud}
\end{figure}

\begin{figure}[t]
    \centering
    \includegraphics[width=\linewidth]{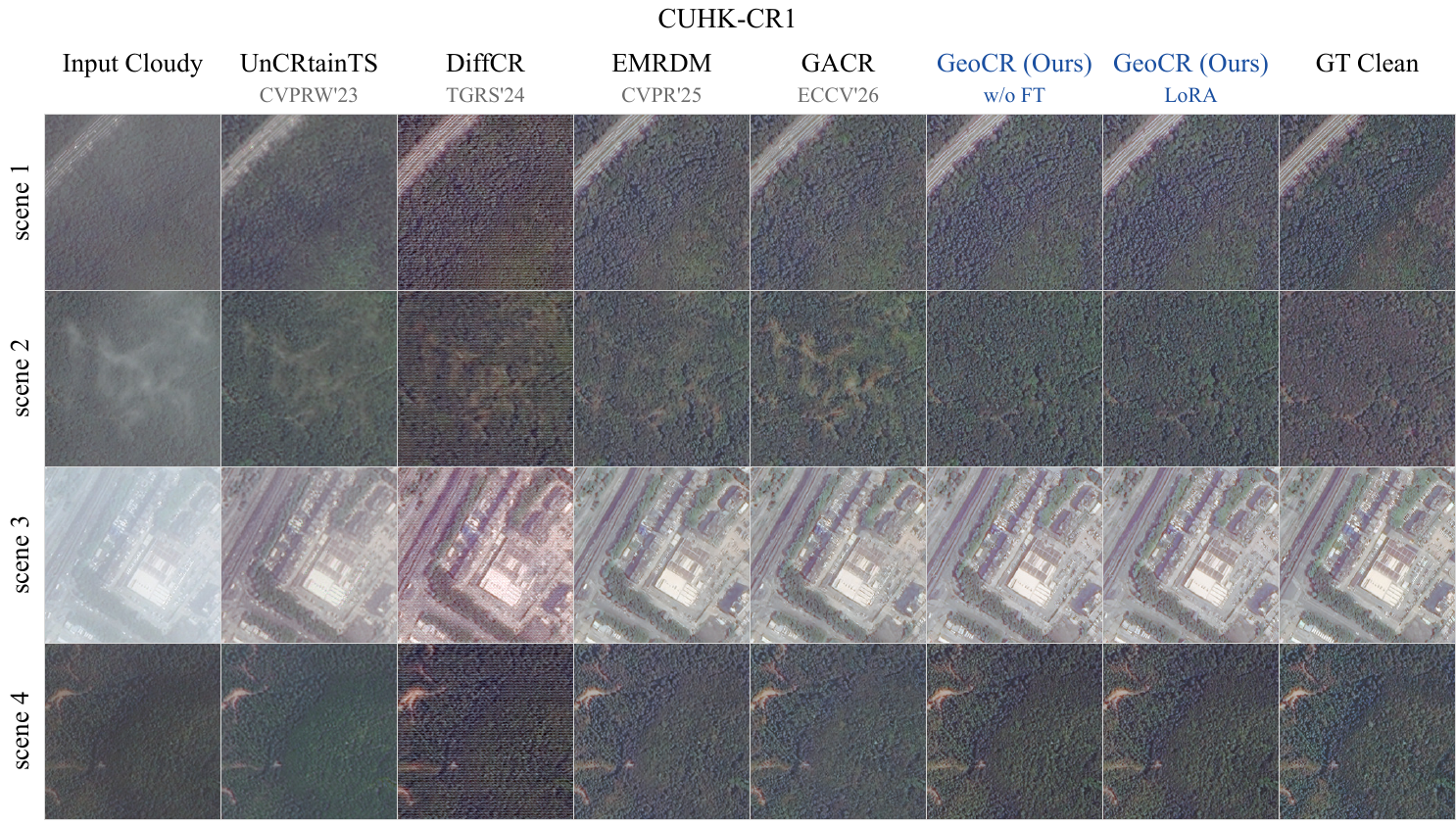}
    \caption{\textbf{Qualitative comparison on CUHK-CR1~\citep{sui2024diffusion}.} Four examples from the RGB+NIR setting, shown as RGB views with paired clear references.
    GeoCR is evaluated without fine-tuning (w/o FT) and with LoRA adaptation.}
    \label{fig:qual_cuhk1}
\end{figure}

\begin{figure}[t]
    \centering
    \includegraphics[width=\linewidth]{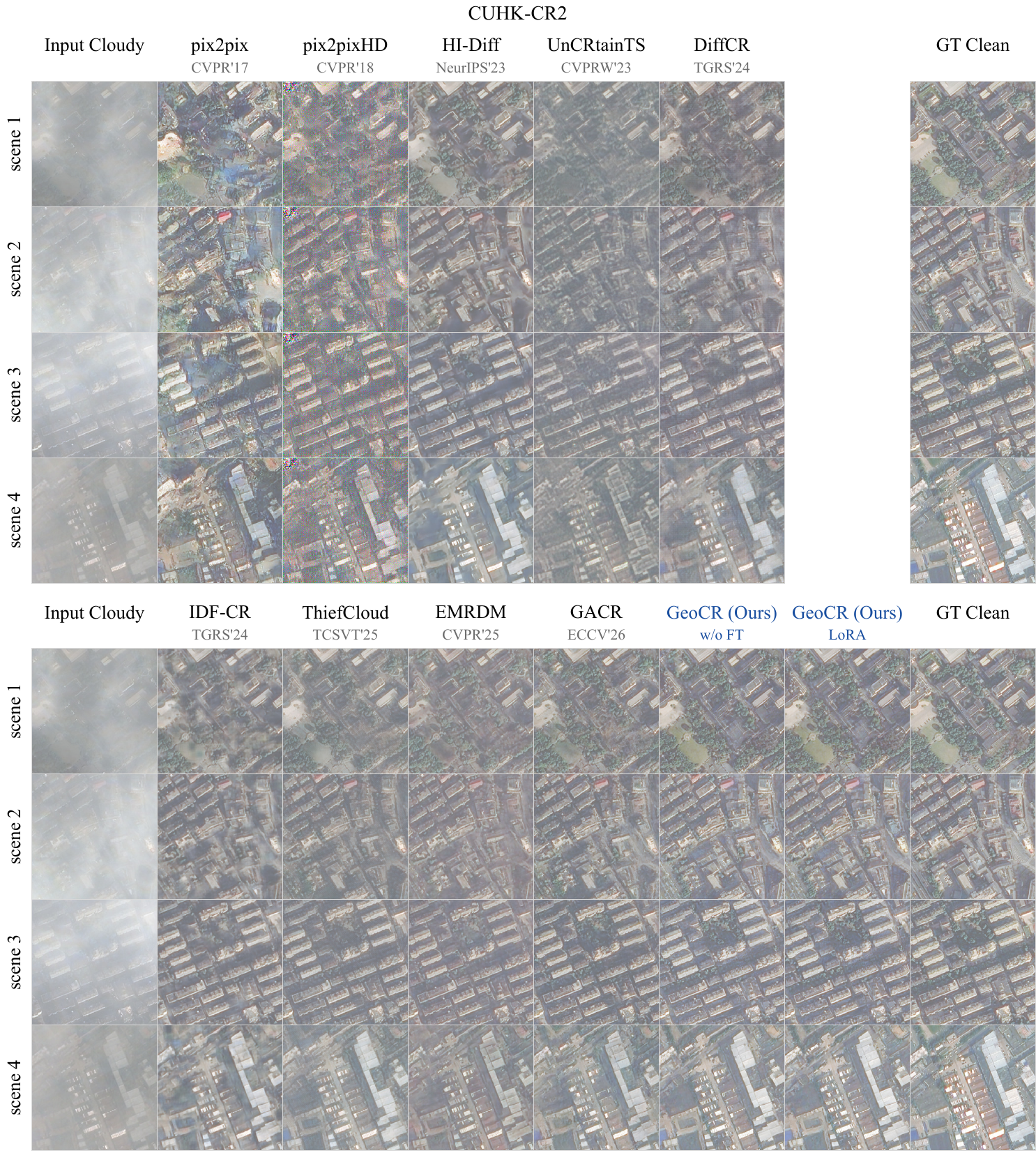}
    \caption{\textbf{Qualitative comparison on CUHK-CR2~\citep{sui2024diffusion}.} The two panels show the same four scenes with different groups of competing methods under the RGB evaluation setting.
    GeoCR is evaluated without fine-tuning (w/o FT) and with LoRA adaptation.}
    \label{fig:qual_cuhk2}
\end{figure}

\begin{figure}[t]
    \centering
    \includegraphics[width=\linewidth]{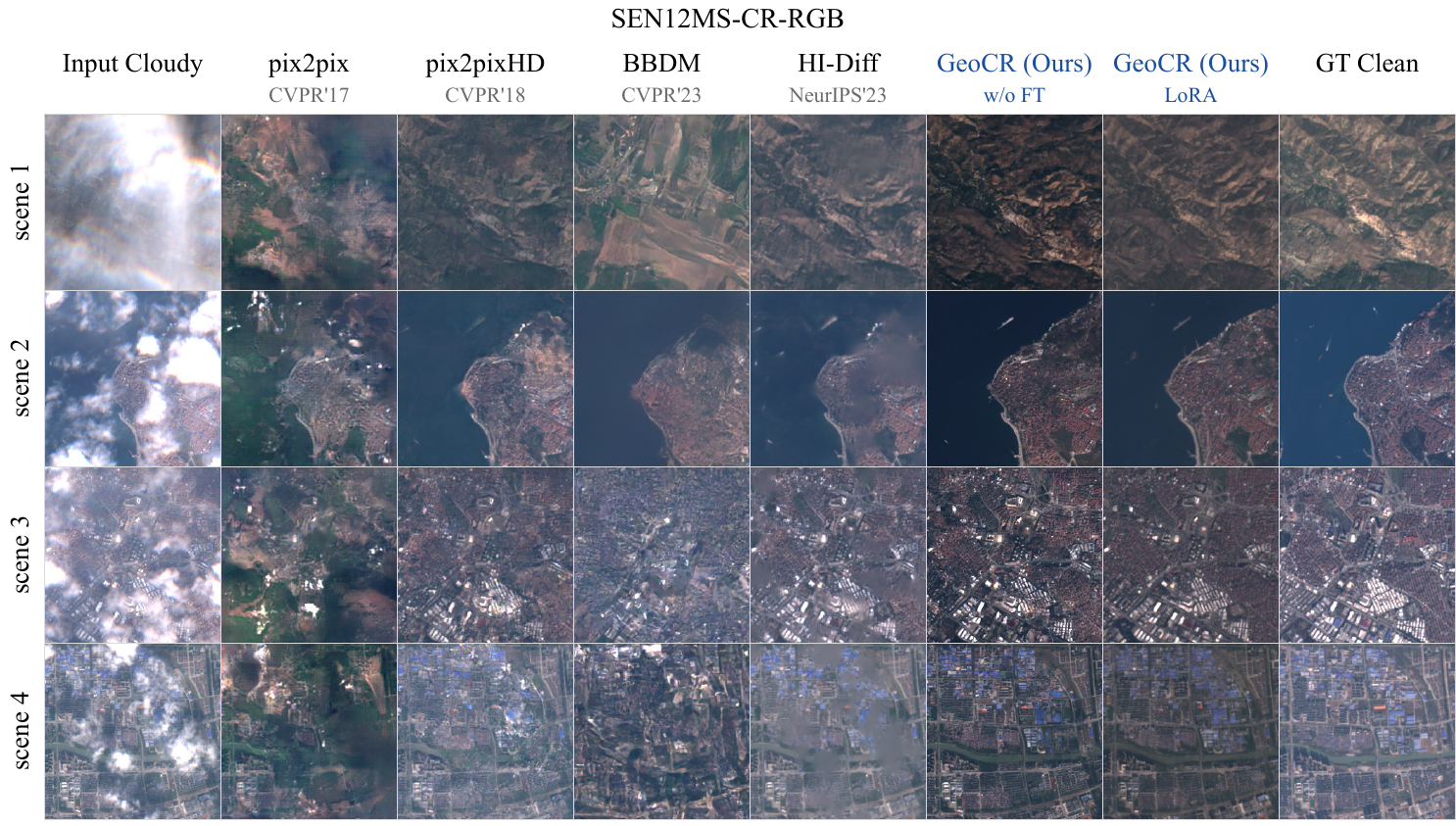}
    \caption{\textbf{Qualitative comparison on SEN12MS-CR~\citep{ebel2021multisensor} (RGB-only).} Four examples from the RGB-only setting, with one cloudy RGB input and no SAR guidance.
    GeoCR is evaluated without fine-tuning (w/o FT) and with LoRA adaptation.}
    \label{fig:qual_sen12mscr_rgb}
\end{figure}

\begin{figure}[t]
    \centering
    \includegraphics[width=\linewidth]{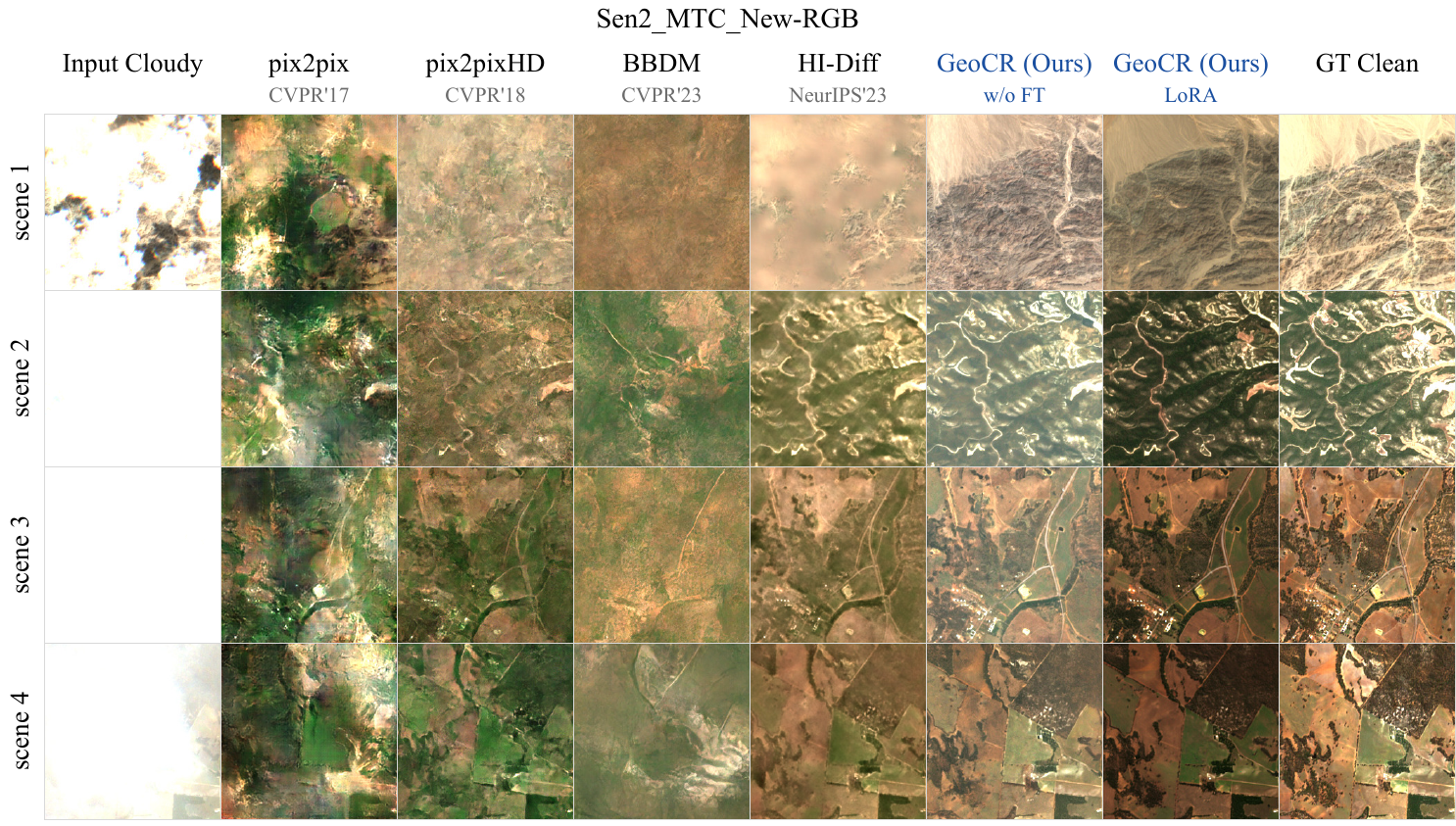}
    \caption{\textbf{Qualitative comparison on Sen2\_MTC\_New~\citep{huang2022ctgan} (RGB-only).} Four examples from the RGB-only setting, with one cloudy RGB input and no SAR guidance.
    GeoCR is evaluated without fine-tuning (w/o FT) and with LoRA adaptation.}
    \label{fig:qual_mtc_new_rgb}
\end{figure}

\begin{figure}[t]
    \centering
    \includegraphics[width=\linewidth]{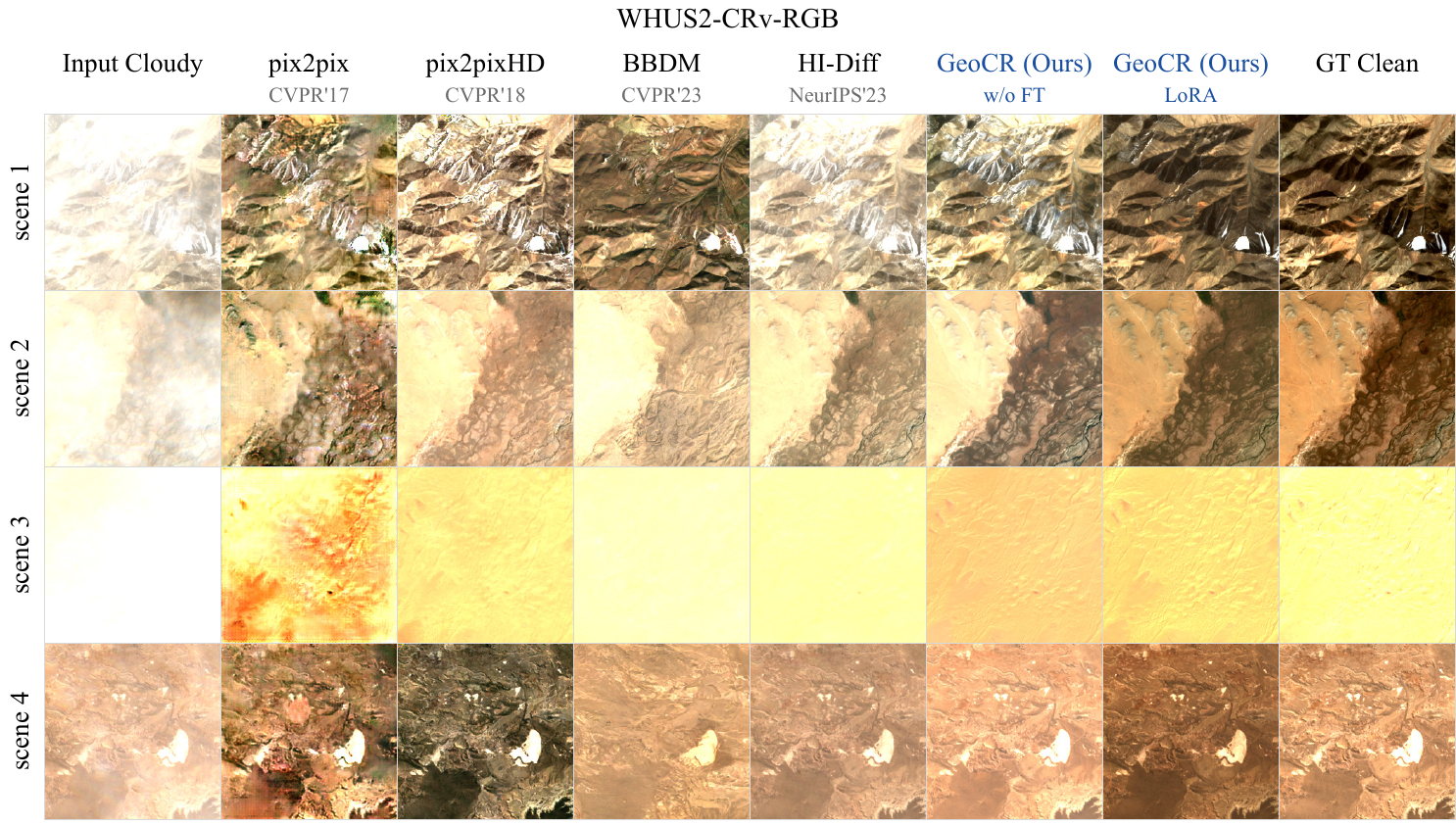}
    \caption{\textbf{Qualitative comparison on WHUS2-CRv~\citep{li2022thin} (RGB-only).} Four examples from the RGB-only setting, with one cloudy RGB input and no SAR guidance.
    GeoCR is evaluated without fine-tuning (w/o FT) and with LoRA adaptation.}
    \label{fig:qual_whus_rgb}
\end{figure}

\paragraph{Additional examples by dataset.}
Figures~\ref{fig:qual_sen12mscr}--\ref{fig:qual_whus_rgb},
collected at the end of this \textit{Appendix}, provide four examples
for each of the six native dataset settings and three RGB-only
settings. Each figure compares GeoCR (w/o FT) and GeoCR (LoRA)
with the available baselines and paired clear references.
All visualizations show RGB views.

\section{Failure Cases and Limitations}
\label{sec:supp_limits}
\begin{figure}[tbp]
    \centering
    \includegraphics[width=0.9\linewidth]{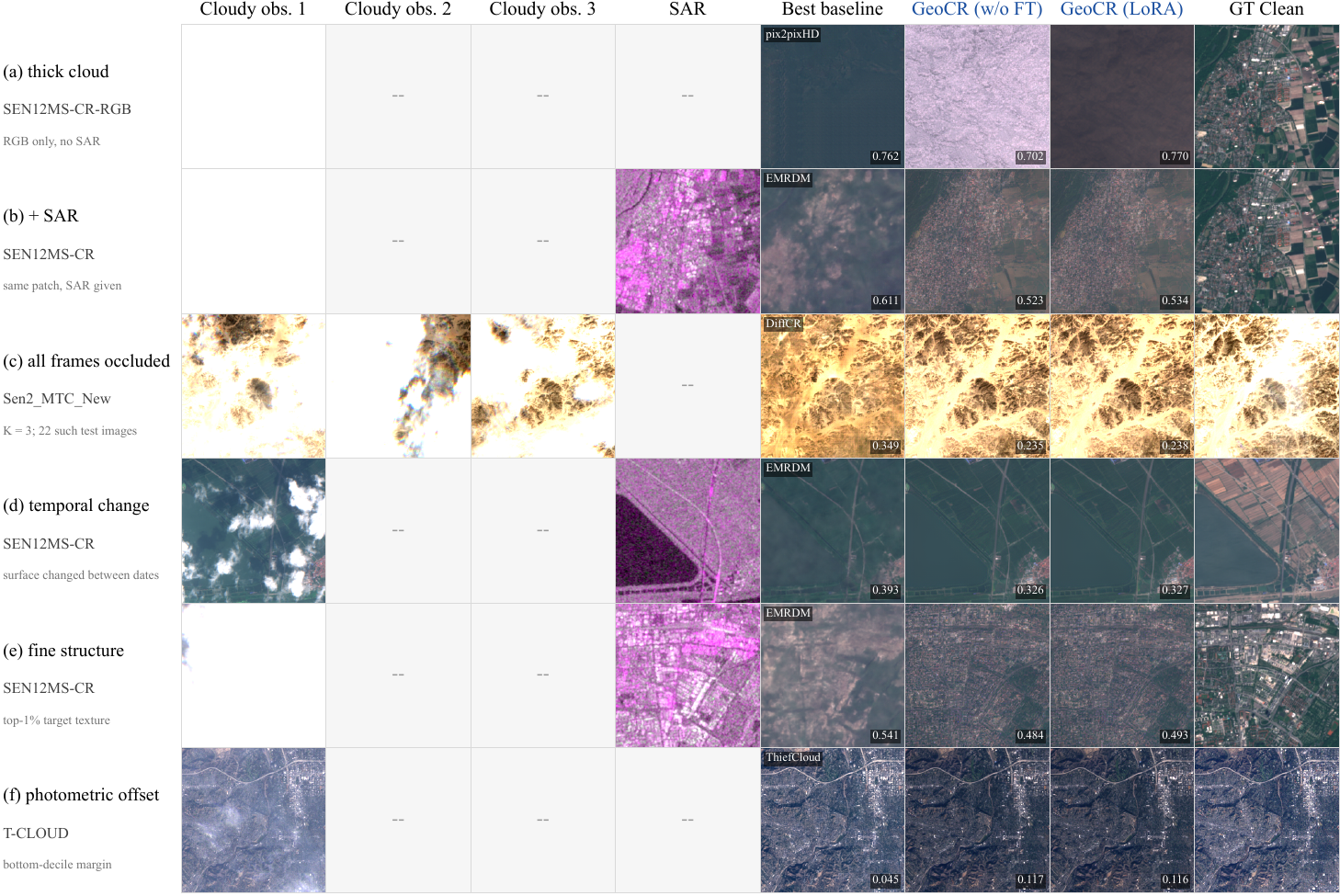}
    \caption{\textbf{Challenging cloud removal cases.}
    Examples cover severe occlusion, the same scene with
    multispectral and SAR inputs, occlusion across all temporal
    observations, temporal surface change, fine urban structure,
    and photometric mismatch.
    Cases are selected using occlusion, change, texture, or error
    criteria. Numbers indicate per-image LPIPS.
    Rows (a) and (b) change both optical bands and SAR availability.}
    \label{fig:supp_hard}
\end{figure}

Figure~\ref{fig:supp_hard} examines cases in which the available
observations or the learned prior are insufficient to reproduce
the reference reliably.

\paragraph{Incomplete evidence.}
When every optical observation is heavily obscured, generated
details may appear plausible while differing from the reference.
SAR provides complementary structural evidence but does not
determine a unique optical appearance.
Fine structures under complete cloud therefore remain difficult
to recover reliably, even when the scene layout is convincing.

\paragraph{Acquisition and radiometric differences.}
Surface changes between acquisitions can make the observed scene
inconsistent with the target. The model may preserve the input
state rather than reconstruct an unobserved state at another time.
Photometric offsets also remain visible on T-CLOUD.
PSNR and SSIM measure agreement with the supplied reference,
while perceptual metrics characterize complementary aspects of
quality; neither establishes the physical accuracy of obscured content.

\paragraph{Scope of generalist behavior.}
GeoCR supports the evaluated spectral, temporal, and SAR
configurations through one pretrained parent.
All evaluated source datasets contribute training data, and
transfer to an entirely unseen corpus is not assessed.
Further evaluation is needed for unseen sensors, geographically
independent transfer, and applications requiring calibrated
spectral fidelity.

\section{Datasets and Preprocessing}
\label{sec:supp_data}

\paragraph{Training corpus and evaluation scope.}
We combine the training splits of ten cloud removal datasets,
totaling 883,331 cloud-free target images
(Table~\ref{tab:supp_sources}).
AllClear, Sen2\_MTC\_Old, and RICE1/2 contribute only pretraining
data; the remaining six sources also provide evaluation benchmarks.
Three additional RGB-only settings are derived from SEN12MS-CR,
Sen2\_MTC\_New, and WHUS2-CRv.
Evaluation uses held-out samples from the contributing sources.

\begin{table}[tbp]
\scriptsize\centering
\caption{\textbf{Dataset representations and pretraining mixture.} Bands describe optical targets; $K$ is the number of cloudy observations. Mix denotes the source sampling probability.}
\label{tab:supp_sources}
\renewcommand{\arraystretch}{1.1}\setlength{\tabcolsep}{4pt}
\resizebox{\linewidth}{!}{\begin{tabular}{lllccr}
\toprule
\rowcolor{blue!5}
\textbf{Source}&\textbf{Optical representation}&\textbf{SAR}&$K$&\textbf{Mix (\%)}&\textbf{Training targets}\\
\midrule
AllClear~\citep{zhou2024allclear}&13 bands, L1C TOA&VV/VH&1--3&53.14&662,022\\
SEN12MS-CR~\citep{ebel2021multisensor}&13 bands, L1C TOA&VV/VH&1&32.93&107,143\\
WHUS2-CRv~\citep{li2022thin}&13 bands, BOA; B10 TOA&--&1&8.81&18,816\\
Sen2\_MTC\_Old~\citep{sarukkai2020cloud}&RGB; NIR in cloudy inputs only&--&3&2.00&88,874\\
Sen2\_MTC\_New~\citep{huang2022ctgan}&RGB+B8, BOA&--&3&1.10&2,380\\
T-CLOUD~\citep{ding2022uncertainty}&RGB, 8-bit&--&1&1.10&2,234\\
RICE2~\citep{lin2019rice}&RGB, 8-bit&--&1&0.27&553\\
CUHK-CR1~\citep{sui2024diffusion}&RGB+NIR, 8-bit&--&1&0.25&508\\
CUHK-CR2~\citep{sui2024diffusion}&RGB+NIR, 8-bit&--&1&0.21&426\\
RICE1~\citep{lin2019rice}&RGB, 8-bit&--&1&0.19&375\\
\midrule
\textbf{Total}&&&&100.00&\textbf{883,331}\\
\bottomrule
\end{tabular}}
\end{table}

\paragraph{Optical and SAR normalization.}
For Sentinel-2 imagery, we convert the stored digital numbers
(DN) to reflectance as $\rho=\mathrm{DN}/10^4$ and normalize
inputs as $2\operatorname{clip}(\rho,0,1)-1$.
Eight-bit imagery is normalized as $2u/255-1$, where $u$
denotes the stored pixel value.
For the separately rendered RGB-only datasets, we first compute
$u=\operatorname{round}[255\operatorname{clip}(2\rho,0,1)]$
and apply the same 8-bit normalization.
SAR measurements $s$ in decibels (dB) are clipped to $[-30,0]$
and mapped to $[-1,1]$ as
$2[\operatorname{clip}(s,-30,0)+30]/30-1$.
TOA and BOA denote top- and bottom-of-atmosphere reflectance,
respectively; L1C denotes the Sentinel-2 Level-1C product.

\paragraph{Spatial and temporal inputs.}
WHUS2-CRv bands are aligned to the 10\,m grid by nearest-neighbor
replication of the 20\,m and 60\,m bands.
For AllClear, cloudy observations are sampled without replacement
within $\pm40$ days of the target, excluding its acquisition date;
SAR is the nearest available acquisition within this window.
Eligible AllClear targets contain at most 10\% cloud, 10\% shadow,
and 1\% no-data coverage.

\paragraph{Split checks.}
We use exact region-of-interest (ROI) identifiers for SEN12MS-CR,
correct scene assignments in WHUS2-CRv, and exclude 104
Sen2\_MTC\_Old samples that overlap Sen2\_MTC\_New test tiles.
CUHK-CR1 evaluation excludes 32 train/validation duplicates and
four within-test duplicates.
RICE1/2 contribute only pretraining data because their supplied
splits contain duplicate images.
We will release the split-audit results documenting these
exclusions.

\section{Evaluation Protocols}
\label{sec:supp_eval}
\label{app:evaluation_metrics}

\paragraph{Test sets and spectral scope.}
Table~\ref{tab:supp_eval} lists the evaluated test splits and the
bands used for PSNR and SSIM.
FID, DISTS, KID, DINO similarity, and LPIPS are computed on RGB
views, including in the native multispectral settings.
Within each setting, competing methods are evaluated on common
samples and spatial extents for each metric.
FID and DISTS are the primary measures of distributional realism
and paired perceptual fidelity, respectively.
Lower values are better for FID, DISTS, KID, and LPIPS;
higher values are better for DINO similarity, SSIM, and PSNR.

\begin{table}[tbp]
\scriptsize\centering
\caption{\textbf{Evaluation settings.} $N_{\mathrm{test}}$ denotes the size of each evaluated test split. CUHK-CR1 counts reflect the duplicate exclusions described in Section~\ref{sec:supp_data}.}
\label{tab:supp_eval}
\renewcommand{\arraystretch}{1.1}\setlength{\tabcolsep}{5pt}
\resizebox{0.5\linewidth}{!}{\begin{tabular}{llr}
\toprule\rowcolor{blue!5}
\textbf{Setting}&\textbf{PSNR/SSIM bands}&$N_{\mathrm{test}}$\\
\midrule
SEN12MS-CR&13 bands&7,899\\
Sen2\_MTC\_New&RGB+B8&687\\
WHUS2-CRv&13 bands&3,746\\
T-CLOUD&RGB&588\\
CUHK-CR1&RGB+NIR&98\\
CUHK-CR2&RGB&111\\
SEN12MS-CR, RGB-only&RGB&7,899\\
Sen2\_MTC\_New, RGB-only&RGB&687\\
WHUS2-CRv, RGB-only&RGB&3,746\\
\bottomrule\end{tabular}}
\end{table}

\paragraph{Distributional measures.}
FID~\citep{heusel2017gans} compares Gaussian approximations to
the generated and reference Inception-feature distributions.
KID~\citep{binkowski2018demystifying} estimates squared maximum
mean discrepancy using a polynomial kernel.
Both use 2,048-dimensional Inception pool-3 features with
$299\times299$ bilinear resizing.
These measures compare image sets and do not directly evaluate
correspondence to an individual cloudy input.

\paragraph{Perceptual and semantic measures.}
DISTS~\citep{ding2020dists} compares structural and textural
statistics using its released VGG-16 model.
LPIPS~\citep{zhang2018lpips} measures distances between normalized
deep features with learned perceptual weights; we use AlexNet v0.1.
DINO similarity~\citep{simeoni2025dinov3} averages cosine similarity
between spatially aligned patch tokens from the frozen
DINOv3-SAT ViT-L/16, excluding CLS and register tokens.
All three compare each reconstruction with its paired clear reference.

\paragraph{Reference fidelity.}
SSIM~\citep{wang2004image} compares local luminance, contrast,
and structure, while PSNR expresses pixelwise mean squared error
relative to the squared intensity range on a logarithmic scale.
We average PSNR over per-image scores.
SSIM uses an $11\times11$ Gaussian window for native Sentinel-2
settings and a $7\times7$ uniform window for display/RGB settings.
These measures quantify spatial and radiometric agreement with
the reference and remain sensitive to misregistration and
acquisition differences.

\clearpage
\clearpage

\clearpage

\clearpage

\clearpage

\clearpage

\clearpage

\clearpage

\clearpage

\clearpage

\clearpage

\bibliography{iclr2027_conference}

@inproceedings{ebel2023uncrtaints,
  title={UnCRtainTS: Uncertainty quantification for cloud removal in optical satellite time series},
  author={Ebel, Patrick and Garnot, Vivien Sainte Fare and Schmitt, Michael and Wegner, Jan Dirk and Zhu, Xiao Xiang},
  booktitle={2023 IEEE/CVF Conference on Computer Vision and Pattern Recognition Workshops (CVPRW)},
  pages={2086--2096},
  year={2023},
  organization={IEEE}
}

@article{zhou2024allclear,
  title={Allclear: A comprehensive dataset and benchmark for cloud removal in satellite imagery},
  author={Zhou, Hangyu and Kao, Chia-Hsiang and Phoo, Cheng Perng and Mall, Utkarsh and Hariharan, Bharath and Bala, Kavita},
  journal={Advances in Neural Information Processing Systems},
  volume={37},
  pages={53571--53597},
  year={2024}
}

@article{zou2024diffcr,
  title={DiffCR: A fast conditional diffusion framework for cloud removal from optical satellite images},
  author={Zou, Xuechao and Li, Kai and Xing, Junliang and Zhang, Yu and Wang, Shiying and Jin, Lei and Tao, Pin},
  journal={IEEE Transactions on Geoscience and Remote Sensing},
  volume={62},
  pages={1--14},
  year={2024},
  publisher={IEEE}
}

@inproceedings{liu2025effective,
  title={Effective cloud removal for remote sensing images by an improved mean-reverting denoising model with elucidated design space},
  author={Liu, Yi and Li, Wengen and Guan, Jihong and Zhou, Shuigeng and Zhang, Yichao},
  booktitle={2025 IEEE/CVF Conference on Computer Vision and Pattern Recognition (CVPR)},
  pages={17851--17861},
  year={2025},
  organization={IEEE}
}

@article{lipman2022flow,
  title={Flow matching for generative modeling},
  author={Lipman, Yaron and Chen, Ricky TQ and Ben-Hamu, Heli and Nickel, Maximilian and Le, Matt},
  journal={arXiv preprint arXiv:2210.02747},
  year={2022}
}

@article{hu2021lora,
  title={Lora: Low-rank adaptation of large language models},
  author={Hu, Edward J and Shen, Yelong and Wallis, Phillip and Allen-Zhu, Zeyuan and Li, Yuanzhi and Wang, Shean and Wang, Lu and Chen, Weizhu},
  journal={arXiv preprint arXiv:2106.09685},
  year={2021}
}

@inproceedings{isola2017image,
  title={Image-to-image translation with conditional adversarial networks},
  author={Isola, Phillip and Zhu, Jun-Yan and Zhou, Tinghui and Efros, Alexei A},
  booktitle={2017 IEEE conference on computer vision and pattern recognition (CVPR)},
  pages={5967--5976},
  year={2017},
  organization={Ieee}
}

@inproceedings{wang2018high,
  title={High-resolution image synthesis and semantic manipulation with conditional gans},
  author={Wang, Ting-Chun and Liu, Ming-Yu and Zhu, Jun-Yan and Tao, Andrew and Kautz, Jan and Catanzaro, Bryan},
  booktitle={2018 IEEE/CVF conference on computer vision and pattern recognition},
  pages={8798--8807},
  year={2018},
  organization={Ieee}
}

@inproceedings{li2023bbdm,
  title={Bbdm: Image-to-image translation with brownian bridge diffusion models},
  author={Li, Bo and Xue, Kaitao and Liu, Bin and Lai, Yu-Kun},
  booktitle={2023 IEEE/CVF Conference on Computer Vision and Pattern Recognition (CVPR)},
  pages={1952--1961},
  year={2023},
  organization={IEEE}
}

@article{chen2023hierarchical,
  title={Hierarchical integration diffusion model for realistic image deblurring},
  author={Chen, Zheng and Zhang, Yulun and Liu, Ding and Gu, Jinjin and Kong, Linghe and Yuan, Xin and others},
  journal={Advances in neural information processing systems},
  volume={36},
  pages={29114--29125},
  year={2023}
}

@article{cui2026llars,
  title={A Unified Foundation Model for All-in-One Multi-Modal Remote Sensing Image Restoration and Fusion with Language Prompting},
  author={Cui, Yongchuan and Liu, Peng},
  journal={arXiv preprint arXiv:2604.05629},
  year={2026}
}

@article{lehmann2026eovae,
  title={EO-VAE: Towards A Multi-sensor Tokenizer for Earth Observation Data},
  author={Lehmann, Nils and Wang, Yi and Xiong, Zhitong and Zhu, Xiaoxiang},
  journal={arXiv preprint arXiv:2602.12177},
  year={2026}
}

@article{zhang2025units,
  title={Units: unified spatio-temporal generative model for remote sensing},
  author={Zhang, Yuxiang and Liang, Shunlin and Li, Wenyuan and Ma, Han and Xu, Jianglei and Ma, Yichuan and Xie, Jiangwei and Li, Wei and Zhang, Mengmeng and Tao, Ran and others},
  journal={arXiv preprint arXiv:2512.04461},
  year={2025}
}

@article{wang2024idfcr,
  title={IDF-CR: Iterative diffusion process for divide-and-conquer cloud removal in remote-sensing images},
  author={Wang, Meilin and Song, Yexing and Wei, Pengxu and Xian, Xiaoyu and Shi, Yukai and Lin, Liang},
  journal={IEEE Transactions on Geoscience and Remote Sensing},
  volume={62},
  pages={1--14},
  year={2024},
  publisher={IEEE}
}

@article{zhao2025thiefcloud,
  title={ThiefCloud: A thickness fused thin cloud removal network for optical remote sensing image with self-supervised learnable cloud prior},
  author={Zhao, Anqi and Feng, Ruitao and Li, Xinghua},
  journal={IEEE Transactions on Circuits and Systems for Video Technology},
  volume={35},
  number={12},
  pages={11834--11848},
  year={2025},
  publisher={IEEE}
}

@inproceedings{wang2026gacr,
  title={Interpretation-Oriented Cloud Removal via Observation-Anchored Residual Flow with Geo-Contextual Alignment},
  author={Wang, Ziyao and Wang, Maonan and He, Yucheng and Ma, Xianping and Wang, Ziyi and Zhang, Hongyang and Cheng, Yirong and Pun, Man-on},
  booktitle={European Conference on Computer Vision},
  pages={248--265},
  year={2026},
  organization={Springer}
}

@article{lin2019rice,
  title={A remote sensing image dataset for cloud removal},
  author={Lin, Daoyu and Xu, Guangluan and Wang, Xiaoke and Wang, Yang and Sun, Xian and Fu, Kun},
  journal={arXiv preprint arXiv:1901.00600},
  year={2019}
}

@article{sui2024diffusion,
  title={Diffusion enhancement for cloud removal in ultra-resolution remote sensing imagery},
  author={Sui, Jialu and Ma, Yiyang and Yang, Wenhan and Zhang, Xiaokang and Pun, Man-On and Liu, Jiaying},
  journal={IEEE Transactions on Geoscience and Remote Sensing},
  volume={62},
  pages={1--14},
  year={2024},
  publisher={IEEE}
}

@article{ebel2021multisensor,
  title={Multisensor data fusion for cloud removal in global and all-season sentinel-2 imagery},
  author={Ebel, Patrick and Meraner, Andrea and Schmitt, Michael and Zhu, Xiao Xiang},
  journal={IEEE Transactions on Geoscience and Remote Sensing},
  volume={59},
  number={7},
  pages={5866--5878},
  year={2020},
  publisher={IEEE}
}

@inproceedings{huang2022ctgan,
  title={Ctgan: Cloud transformer generative adversarial network},
  author={Huang, Gi-Luen and Wu, Pei-Yuan},
  booktitle={2022 IEEE international conference on image processing (ICIP)},
  pages={511--515},
  year={2022},
  organization={IEEE}
}

@article{li2022thin,
  title={Thin cloud removal fusing full spectral and spatial features for Sentinel-2 imagery},
  author={Li, Jun and Zhang, Yuejie and Sheng, Qinghong and Wu, Zhaocong and Wang, Bo and Hu, Zhongwen and Shen, Guanting and Schmitt, Michael and Molinier, Matthieu},
  journal={IEEE Journal of Selected Topics in Applied Earth Observations and Remote Sensing},
  volume={15},
  pages={8759--8775},
  year={2022},
  publisher={IEEE}
}

@inproceedings{sarukkai2020cloud,
  title={Cloud removal in satellite images using spatiotemporal generative networks},
  author={Sarukkai, Vishnu and Jain, Anirudh and Uzkent, Burak and Ermon, Stefano},
  booktitle={2020 IEEE Winter Conference on Applications of Computer Vision (WACV)},
  pages={1785--1794},
  year={2020},
  organization={IEEE}
}

@inproceedings{ding2022uncertainty,
  title={Uncertainty-based thin cloud removal network via conditional variational autoencoders},
  author={Ding, Haidong and Zi, Yue and Xie, Fengying},
  booktitle={Asian Conference on Computer Vision},
  pages={52--68},
  year={2022},
  organization={Springer}
}

@misc{flux-2-2025,
    author={Black Forest Labs},
    title={{FLUX.2: Frontier Visual Intelligence}},
    year={2025},
    howpublished={\url{https://bfl.ai/blog/flux-2}},
}

@article{heusel2017gans,
  title={Gans trained by a two time-scale update rule converge to a local nash equilibrium},
  author={Heusel, Martin and Ramsauer, Hubert and Unterthiner, Thomas and Nessler, Bernhard and Hochreiter, Sepp},
  journal={Advances in neural information processing systems},
  volume={30},
  year={2017}
}

@article{ding2020dists,
  title={Image quality assessment: Unifying structure and texture similarity},
  author={Ding, Keyan and Ma, Kede and Wang, Shiqi and Simoncelli, Eero P},
  journal={IEEE transactions on pattern analysis and machine intelligence},
  volume={44},
  number={5},
  pages={2567--2581},
  year={2020},
  publisher={IEEE}
}

@inproceedings{zhang2018lpips,
  title={The unreasonable effectiveness of deep features as a perceptual metric},
  author={Zhang, Richard and Isola, Phillip and Efros, Alexei A and Shechtman, Eli and Wang, Oliver},
  booktitle={2018 IEEE/CVF conference on computer vision and pattern recognition},
  pages={586--595},
  year={2018},
  organization={IEEE}
}

@article{wang2004image,
  title={Image quality assessment: from error visibility to structural similarity},
  author={Wang, Zhou and Bovik, Alan C and Sheikh, Hamid R and Simoncelli, Eero P},
  journal={IEEE transactions on image processing},
  volume={13},
  number={4},
  pages={600--612},
  year={2004},
  publisher={IEEE}
}

@article{binkowski2018demystifying,
  title={Demystifying mmd gans},
  author={Bi{\'n}kowski, Miko{\l}aj and Sutherland, Danica J and Arbel, Michael and Gretton, Arthur},
  journal={arXiv preprint arXiv:1801.01401},
  year={2018}
}

@article{simeoni2025dinov3,
  title={Dinov3},
  author={Sim{\'e}oni, Oriane and Vo, Huy V and Seitzer, Maximilian and Baldassarre, Federico and Oquab, Maxime and Jose, Cijo and Khalidov, Vasil and Szafraniec, Marc and Yi, Seungeun and Ramamonjisoa, Micha{\"e}l and others},
  journal={arXiv preprint arXiv:2508.10104},
  year={2025}
}

@inproceedings{blau2018perception,
  title={The perception-distortion tradeoff},
  author={Blau, Yochai and Michaeli, Tomer},
  booktitle={2018 IEEE/CVF Conference on Computer Vision and Pattern Recognition},
  pages={6228--6237},
  year={2018},
  organization={IEEE}
}
\bibliographystyle{iclr2027_conference}

\end{document}